\pdfoutput=1
\documentclass[journal]{IEEEtran}

\usepackage{microtype}
\usepackage{graphicx}
\usepackage{subfigure}
\usepackage{booktabs} 
\usepackage{hyperref}
\usepackage{algorithm}
\usepackage{algpseudocode}
\usepackage{balance} 
\usepackage{graphicx}
\usepackage{subfigure}
\usepackage{float}
\usepackage{amsfonts}
\usepackage{enumerate}
\usepackage{amsmath}
\usepackage{enumitem}
\usepackage[marginal]{footmisc}
\usepackage{caption}

\begin{document}

\title{Multi-agent Deep Covering Skill Discovery}
\author{Jiayu Chen, Marina Haliem, Tian Lan, and Vaneet Aggarwal \thanks{J. Chen, M. Haliem, and V. Aggarwal are with Purdue University, West Lafayette IN 47907, USA, email: \{chen3686,mwadea,vaneet\}@purdue.edu.} 
\thanks{T. Lan is with the George Washington University, Washinton DC 20052, USA, email:tlan@gwu.edu.}
\thanks{This paper was presented in part at the ICML workshop, July 2021 (no proceedings).}}

\maketitle

\begin{abstract}
The use of skills (a.k.a., options) can greatly accelerate exploration in reinforcement learning, especially when only sparse reward signals are available. While option discovery methods have been proposed for individual agents, in multi-agent reinforcement learning settings, discovering collaborative options that can coordinate the behavior of multiple agents and encourage them to visit the under-explored regions of their joint state space has not been considered. In this case, we propose \textit{Multi-agent Deep Covering Option Discovery}, which constructs the multi-agent options through minimizing the expected cover time of the multiple agents' joint state space. 

Also, we propose a novel framework to adopt the multi-agent options in the MARL process. In practice, a multi-agent task can usually be divided into some sub-tasks, each of which can be completed by a sub-group of the agents. Therefore, our algorithm framework first leverages an attention mechanism to find collaborative agent sub-groups that would benefit most from coordinated actions. Then, a hierarchical algorithm, namely \textbf{HA-MSAC}, is developed to learn the multi-agent options for each sub-group to complete their sub-tasks first, and then to integrate them through a high-level policy as the solution of the whole task. This hierarchical option construction allows our framework to strike a balance between scalability and effective collaboration among the agents. 

The evaluation based on multi-agent collaborative tasks shows that the proposed algorithm can effectively capture the agent interactions with the attention mechanism, successfully identify multi-agent options, and significantly outperforms prior works using single-agent options or no options, in terms of both faster exploration and higher task rewards.

\end{abstract}

\begin{IEEEkeywords}Multi-agent Reinforcement Learning, Skill Discovery, Deep Covering Options
\end{IEEEkeywords}

\section{Introduction}\label{intro}

Option discovery \cite{DBLP:journals/ai/SuttonPS99} enables temporally-abstract actions to be constructed in the reinforcement learning process. It can greatly improve the performance of reinforcement learning agents by representing actions at different time scales. 
Among recent developments on the topic, \textit{Covering Option Discovery} \cite{DBLP:conf/icml/JinnaiPAK19} has been shown to be a promising approach. It leverages Laplacian matrix extracted from the state-transition graph induced by the dynamics of the environment. To be specific, the second smallest eigenvalue of the Laplacian matrix, known as the algebraic connectivity of the graph, is considered as a measure of how well-connected the graph is \cite{DBLP:conf/cdc/GhoshB06}. In this case, it uses the algebraic connectivity as an intrinsic reward to train the option policy, with the goal of connecting the states that are not well-connected, encouraging the agent to explore infrequently-visited regions, and thus minimizing the agent's expected cover time of the state space. Recently, deep learning techniques have been developed to extend the use of covering options to large/infinite state space, e.g., \textit{Deep Covering Option Discovery} \cite{DBLP:conf/iclr/JinnaiPMK20}. However, these efforts focus on discovering options for individual agents. Discovering collaborative options that encourage multiple agents to visit the under-explored regions of their joint state space has not been considered. 

In this paper, we propose a novel framework -- \textit{Multi-agent Deep Covering Option Discovery}.
For multi-agent scenarios, recent works \cite{DBLP:conf/atal/ChakravortyWRCB20, DBLP:conf/iclr/LeeYL20, DBLP:conf/atal/YangBZ20} compute options with exploratory behaviors for each individual agent by considering only its own state transitions, and then learn to collaboratively leverage these individual options. However, our proposed framework directly recognize joint options composed of multiple agents’ temporally-abstract action sequences to encourage joint exploration. Also, we note that in practical scenarios, multi-agent tasks can often be divided into a series of sub-tasks and each sub-task can be completed by a sub-group of the agents. Thus, our proposed algorithm leverages an attention mechanism  \cite{DBLP:conf/nips/VaswaniSPUJGKP17} in the option discovery process to quantify the strength of agent interactions and find collaborative agent sub-groups. After that, we can train a set of multi-agent options for each sub-group to complete their sub-tasks, and then integrate them through a high-level policy as the solution for completing the 
whole task. This sub-group partitioning and hierarchical learning structure can effectively construct collaborative options that jointly coordinate the exploration behavior of multiple agents, while keeping the algorithm scalable in practice.

The main contributions of our work are as follows: (1) We extend the deep covering option discovery to a multi-agent scenario, namely \textit{Multi-agent Deep Covering Option Discovery}, and demonstrate that the use of multi-agent options can further improve the performance of MARL agents compared with single-agent options. (2) We propose to leverage an attention mechanism in the discovery process to enable agents to find peer agents that it should interact closely and form sub-groups with. (3)  We propose \textbf{HA-MSAC}, a hierarchical MARL algorithm, which integrates the training of intra-option policies (for the option construction) and the high-level policy (for integrating the options). The proposed algorithm, evaluated on MARL collaborative tasks, significantly outperforms prior works in terms of faster exploration and higher task rewards.



The rest of this paper is organized as follows. Section \ref{related} introduces some related works and highlights the innovation of this paper. Section \ref{background} presents the background knowledge on option discovery and attention mechanism. Section \ref{approach} and \ref{HA-MSAC} explain the proposed approach in detail, including its overall framework, network structure and objective functions to optimize. Section \ref{eval} describes the simulation setup, and presents the comparisons of our algorithm with two baselines: MARL without option discovery, and MARL with single-agent option discovery. Section \ref{conc} concludes this paper.

\section{Related Work} \label{related}

\textbf{Option Discovery.} The option framework was proposed in \cite{DBLP:journals/ai/SuttonPS99}, which extends the usual notion of actions to include options — closed-loop policies for taking actions over a period of time. Formally, a set of options defined over an MDP constitutes a semi-MDP (SMDP), where the SMDP actions (options) are no longer black boxes, but policies in the base MDP which can be learned in their own right. In literature, lots of option discovery algorithms utilize the task-dependent reward signals generated by the environment, such as \cite{menache2002q, konidaris2009skill,harb2018waiting, tiwari2019natural}. Specifically, they directly define or learn through gradient descent the options that can lead the agent to the rewarding states in the environments, and then utilize these trajectory segments (options) to compose the completed trajectory toward the goal state. However, these methods rely on dense reward signals, which are usually hard to acquire in real-life tasks. Therefore, the authors in \cite{DBLP:conf/iclr/EysenbachGIL19} proposed an approach to generate options through maximizing an information theoretic objective so that each option can generate diverse behaviors. It learns useful skills/options without reward signals and thus can be applied in environments where only sparse rewards are available.

On the other hand, the work in \cite{DBLP:conf/icml/MachadoBB17, DBLP:conf/icml/JinnaiPAK19} focused on \textit{Covering Option Discovery}, a method which is also not based on the task-dependent reward signals but on the Laplacian matrix of the environment's state-transition graph. This method aims at minimizing the agent's expected cover time of the state space with a uniformly random policy. To realize this, it augments the agent’s action set with options obtained from the eigenvector associated with the second smallest eigenvalue (algebraic connectivity) of the Laplacian matrix. 
However, this Laplacian-based framework can only be applied to tabular settings. To mitigate this issue, the authors in \cite{DBLP:conf/iclr/JinnaiPMK20} proposed \textit{Deep Covering Option Discovery} to combine covering options with modern representation learning techniques for the eigenfunction estimation, which can be applied in domains with infinite state space. In \cite{DBLP:conf/iclr/JinnaiPMK20}, the authors compared their approach with the one proposed by \cite{DBLP:conf/iclr/EysenbachGIL19} (mentioned above): both approaches are sample-based and scalable to large-scale state space, but RL agents with deep covering options have better performance on the same benchmarks. Thus, in the evaluation part, we use \textit{Deep Covering Option Discovery} as one of the baselines.

Note that all the approaches mentioned above are for single-agent scenarios and the goal of this paper is to extend the adoption of deep covering options to multi-agent reinforcement learning.

\textbf{Options in multi-agent scenarios.} As mentioned in Section \ref{intro}, most of the researches about adopting options in MARL tried to define or learn the options for each individual agent first, and then learn the collaborative behaviors among the agents based on their extended action sets -- \{primitive actions, individual options\}. Therefore, the options they use are still single-agent options, and the coordination in the multi-agent system can only be shown/utilized in the option-choosing process while not the option discovery process. We can classify these works by the option discovery methods they used: the algorithms in \cite{DBLP:conf/atal/AmatoKK14,amato2019modeling} directly defined the options based on their task without the learning process; the algorithms in \cite{shen2006multi, DBLP:conf/atal/ChakravortyWRCB20, DBLP:conf/iclr/LeeYL20} learned the options based on the task-related reward signals generated by the environment; the algorithm in \cite{DBLP:conf/atal/YangBZ20} trained the options based on a reward function that is a weighted summation of the environment reward and the information theoretic reward term proposed in \cite{DBLP:conf/iclr/EysenbachGIL19}.

In this paper, we propose to construct multi-agent deep covering options using the Laplacian-based framework mentioned above.
Also, in an N-agent system, there may be not only N-agent options, but also one-agent options, two-agent options, etc. In this case, we divide the agents into some sub-groups based on their interaction relationship first, which is realized through the attention mechanism \cite{DBLP:conf/nips/VaswaniSPUJGKP17}, and then construct the interaction patterns (multi-agent options) for each sub-group accordingly. Through these improvements, the coordination among agents is considered in the option discovery process, which has the potential to further improve the performance of MARL agents.

\textbf{Hierarchical Multi-agent Reinforcement Learning.} Multi-agent reinforcement learning (MARL) methods hold great potential to solve a variety of real-world problems. Specific to the multi-agent cooperative setting (our focus), there are many related works: VDN \cite{sunehag2018value}, QMIX \cite{rashid2018qmix}, QTRAN \cite{son2019qtran}, MAVEN \cite{mahajan2019maven} and MSAC \cite{DBLP:conf/icml/IqbalS19}. Among them, MSAC introduces the attention mechanism \cite{DBLP:conf/nips/VaswaniSPUJGKP17} to MARL, which is also what we need. 

On the other hand, when adopting options in reinforcement learning, agents need to learn the internal policy of each option (low-level policy) and the policy to choose over the options (high-level policy) in the meantime. Also, the high-level policy is used to select a new option only when the previous option terminates, so the termination signals should be considered when updating the high-level policy. The updating rules of reinforcement learning with options are talked about in literatures like \cite{bacon2017option, DBLP:journals/corr/abs-1712-04065, tiwari2019natural} as the option-critic framework.

In this paper, we try to introduce multi-agent options to MARL, so we extend the option-critic framework to multi-agent scenarios and combine it with MSAC to propose \textbf{HA-MSAC}, a hierarchical multi-agent reinforcement learning algorithm, which will be talked about in Section \ref{HA-MSAC}.

\section{Background} \label{background}

Before extending deep covering options to the multi-agent setting, we will introduce the formal definition of the option framework and some key issues of deep covering options. Also, we will introduce the soft-attention mechanism leveraged for sub-group division in this section.

\subsection{Formal Definition of Option} 

 In this paper, we use the term \textit{options} for the generalization of primitive actions to include temporally-extended courses of actions. As defined in \cite{DBLP:journals/ai/SuttonPS99}, an option $\omega$ consists of three components: an intra-option policy $\pi_{\omega}: \mathcal{S} \text{ x } \mathcal{A} \rightarrow [0,1]$, a termination condition $ \beta_{\omega}: \mathcal{S} \rightarrow \{0,1\}$, and an initiation set $I_{\omega} \subseteq \mathcal{S}$. An option $<I_{\omega}, \pi_{\omega}, \beta_{\omega}>$ is available in state $s$ if and only if $s \in I_{\omega}$. If the option $\omega$ is taken, actions are selected according to $\pi_{\omega}$ until $\omega$ terminates stochastically according to $\beta_{\omega}$. Therefore, in order to get an option, we need to train/define the intra-option policy, and to define the termination condition and initiation set.

\subsection{Deep Covering Option Discovery} \label{dcod}

As descried in \cite{DBLP:conf/iclr/JinnaiPMK20}, deep covering options can be constructed through greedily maximizing the state-space graph's algebraic connectivity -- the second smallest eigenvalue, so as to minimize the expected cover time of the state space. To realize this, they first compute the eigenfunction $f$ associated with the algebraic connectivity by minimizing $G(f)$:

\begin{equation} \label{equ:20}
\begin{aligned}
    G(f) &= \frac{1}{2}\mathbb{E}_{(s, s')\sim \mathcal{H}}[(f(s)-f(s'))^{2}] + \eta\mathbb{E}_{s\sim\rho,s'\sim\rho}\\
    &[(f(s)^2-1)(f(s')^2-1)+2f(s)f(s')]
\end{aligned}
\end{equation}
where $\mathcal{H}$ is the set of sampled state-transitions and $\rho$ is the distribution of the states in $\mathcal{H}$. Note that this is a sample-based approach and thus can scale to infinite state-space domains. Then, based on the computed $f$, they define the termination set as a set of states where the $f$ value is smaller than the $k$-th percentile of the $f$ values on $\mathcal{H}$. Accordingly, the initiation set is defined as the complement of the termination set. As for the intra-option policy, they train it through maximizing the reward, $r(s, a, s') = f(s) - f(s')$, to encourage the agent to explore the states with lower $f$ values, i.e., the less-explored states in the termination set.

In this paper, we will compute $f$ of the joint observation space of each collaborative group of agents, and then learn the multi-agent options based on $f$, so as to encourage the joint exploration of the agents within the same group and to increase the connectivity of their joint observation space.

\subsection{Soft Attention Mechanism}

The soft attention mechanism functions in a manner similar with a differentiable key-value memory model \cite{DBLP:conf/nips/VaswaniSPUJGKP17, DBLP:journals/corr/GravesWD14, DBLP:conf/icml/OhCSL16}. The soft attention weight of agent $i$ to agent $j$ is defined as Equation (\ref{equ:21}):

\begin{equation} \label{equ:21}
\begin{aligned}
    W_{S}^{i,j} = \frac{exp[h_{j}^{T}W_{k}^{T}W_{q}h_{i}]}{\sum_{n \neq i}exp[h_{n}^{T}W_{k}^{T}W_{q}h_{i}]}
\end{aligned}
\end{equation}
where $h_{i} = g_{i}(o_{i})$ is the embedding of agent $i$'s input (we take the observation as the input and a neural network defined in Section \ref{network} as the embedding function), $W_{q}$ transforms $h_{i}$ into a “query” and $W_{k}$ transforms $h_{j}$ into a “key”. In this case, the attention weight $W_{S}^{i,j}$ is acquired through comparing the embedding $h_{j}$ with $h_{i}$ with a bilinear mapping (i.e., the query-key
system) and passing the similarity value between these two
embeddings into a softmax function. Note that the weight matrices for extracting selectors ($W_{q}$), keys ($W_{k}$) are shared across all agents, which encourages a common embedding space.

\begin{figure}[t]
	\centering
	\includegraphics[height=1.8in, width=3.5in]{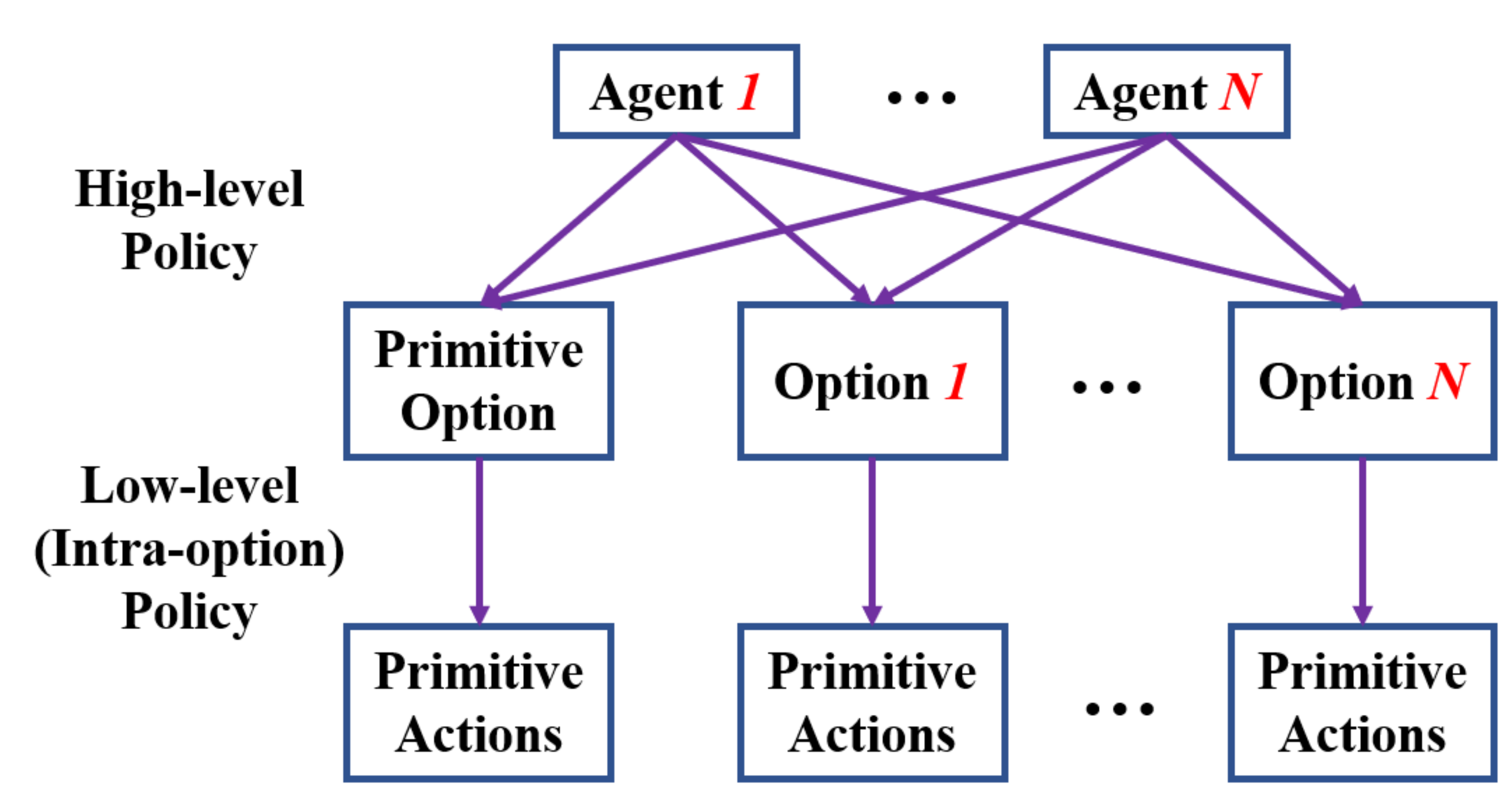}
	\caption{An agent first decides on which option $\omega$ to use according to the high-level policy, and then decides on the action (primitive action) to take based on the corresponding intra-option policy $\pi_{\omega}$. \textbf{Primitive option}: Typically, we train a RL agent to select among the primitive actions; we view this agent as a special option, whose intra-option policy lasts for only one step. \textbf{Option} $\mathbf{1 \sim N}$: Based on the attention mechanism, each agent can figure out which agents to collaborate closely and form a sub-group with, so there are at most $N$ sub-groups (duplicate ones need to be eliminated), and we need to train a multi-agent option for each sub-group.}
	\label{fig:0}
\end{figure}

\begin{figure*}[t]
\centering
\subfigure[Policy Network]{
\label{fig:1(a)} 
\includegraphics[width=3.5in, height=3.5in]{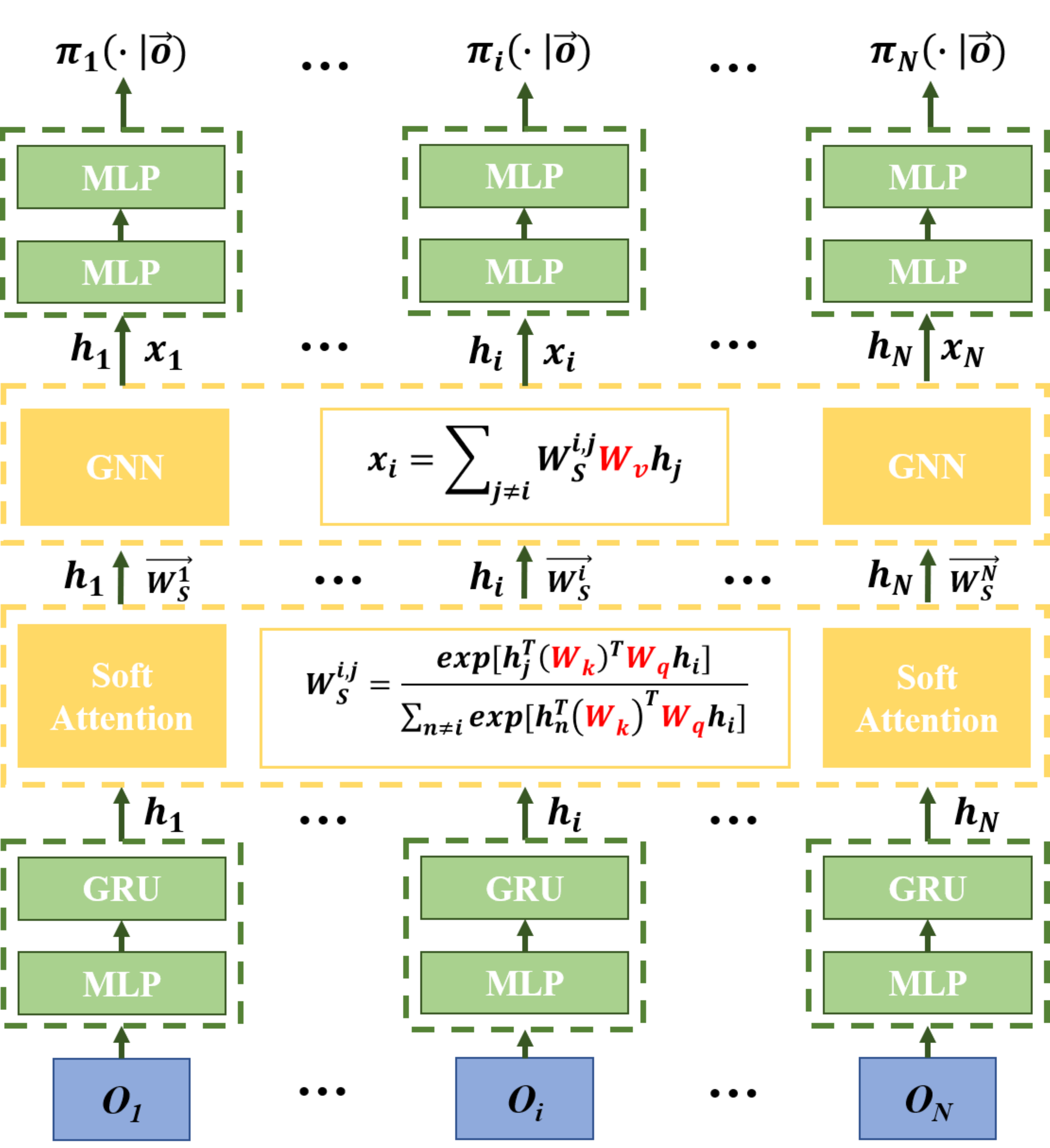}}
\subfigure[Q-function Network]{
\label{fig:1(b)} 
\includegraphics[width=3.5in, height=3.5in]{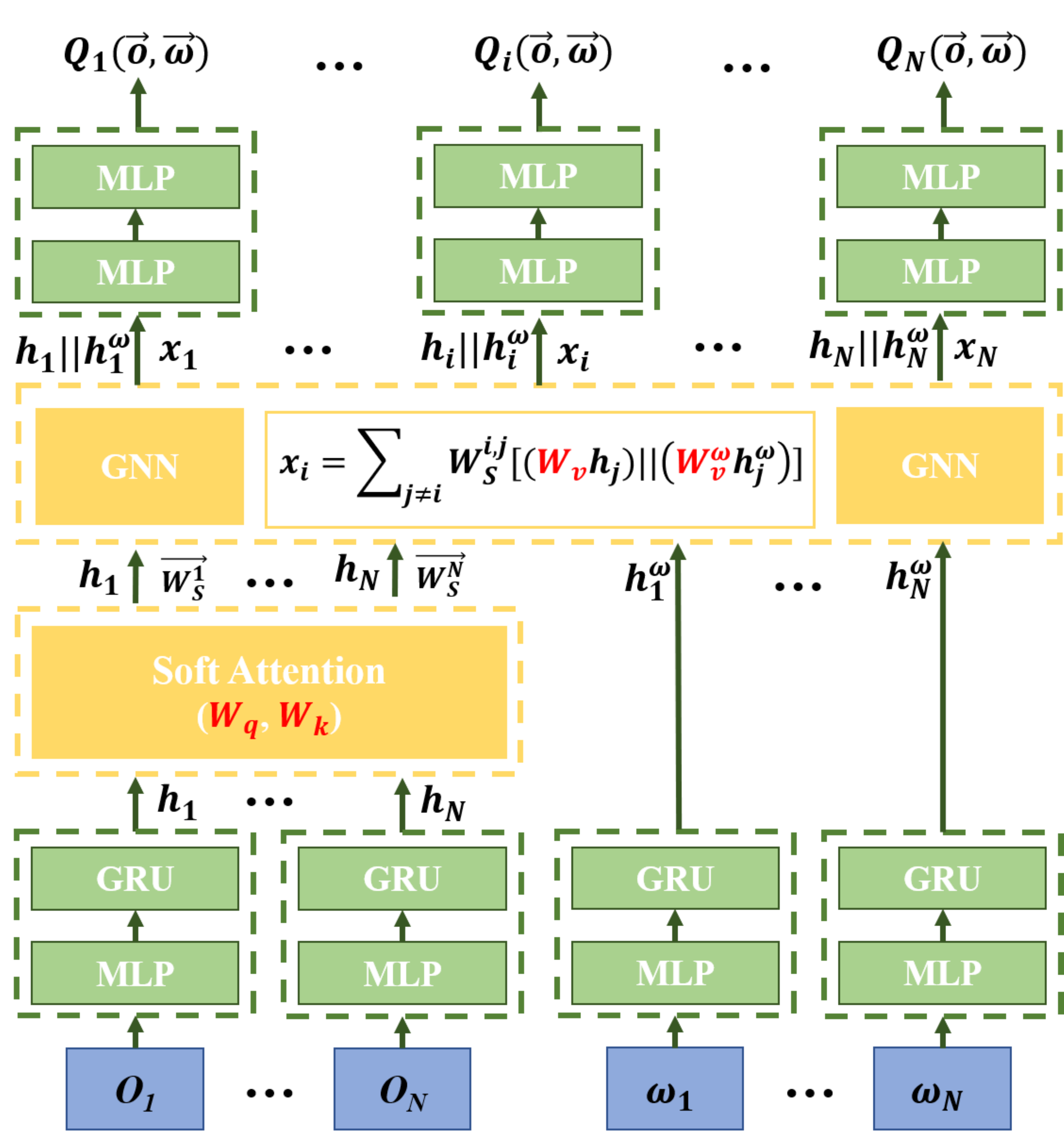}}

\caption{(a) Policy Network: The policy function takes the observation as input and outputs the probability distribution over the action (option) set. The green modules in the figure are specific to each individual agent, while the orange ones are shared by all the agents. (b) Q-function Network: The Q-function takes the observation and action (option) as input and outputs the Q-value for each agent. Note that the observation embedding layers and soft attention modules are shared between the policy network and Q-function network, in order to make sure the interaction relationship extracted by the two networks are the same. (PS: MLP -- Multilayer Perceptron, GRU -- Gated Recurrent Unit \cite{DBLP:journals/corr/ChungGCB14}, GNN -- Graph Neural Network.)}

\label{fig:1} 
\end{figure*}

\section{Algorithm Framework and Network Structure} \label{approach}

In this section, we will introduce \textit{Multi-agent Deep Covering Option Discovery} and how to adopt it in a MARL setting. First, we will provide the key objective functions of \textit{Multi-agent Deep Covering Option Discovery} and the hierarchical algorithm framework to adopt it in MARL. Then, as an important algorithm module, we will show how to integrate the attention mechanism in the network design and how to adopt it for the sub-group division. 

\subsection{Main Framework} \label{framework}

\begin{algorithm}[t]
\caption{Main Framework}
\label{alg:Algo1}
{\begin{algorithmic}[1]
	\State \textbf{Input}: primitive option $A$, high-level policies for each agent $\pi_{1:N}$ and corresponding Q-functions $Q_{1:N}$, generation times of options $N_{\omega}$, generation frequency $N_{int}$;
	\State \textbf{Initialize} the set of options $\Omega' \leftarrow \{A\}$
	\State \textbf{Create} an empty replay buffer $B$
	\State \textbf{Set} $n_{\omega} \leftarrow 0$
	\For {episode $i=1$ to $N_{epi}$}
	    \State \textbf{Collect} a trajectory $\tau_{i}$ by repeating this process until 
	    \Statex $\quad\ $ done: choose an available option from $\Omega'$ according 
	    \Statex $\quad\ $ to $\pi_{1:N}$, and then execute the corresponding intra-
	    \Statex $\quad\ $ option policy until it terminates
	    \State \textbf{Update} $B$ with $\tau_{i}$
	    \If{$i\ mod\ N_{int}==0$ and $n_{\omega}<N_{\omega}$}
	        \State \textbf{Generate} a set of multi-agent options $\Omega$ using 
	        \Statex $\quad\quad\ \ $ Algorithm \ref{alg:Algo2} based on trajectories in $B$ 
	        \State \textbf{Update} $\Omega'$: $\Omega' \leftarrow \{A\} \cup \Omega$
	        \State \textbf{Update} $n_{\omega}$: $n_{\omega} \leftarrow n_{\omega} + 1$
	    \EndIf
	    \State \textbf{Sample} trajectories $\tau_{1:batchsize}$ from $B$
	    \State \textbf{Update} $\pi_{1:N}$, $Q_{1:N}$ using \textbf{HA-MSAC} (defined in Sec
	    \Statex $\quad\ \ $tion \ref{HA-MSAC}) based on $\tau_{1:batchsize}$
	\EndFor
	
\end{algorithmic}}
\end{algorithm}

\begin{algorithm}[t]
\caption{Multi-agent Deep Covering Option Discovery}
\label{alg:Algo2}
{\begin{algorithmic}[1]
    \State \textbf{Input}:  percentile $0 \leq k \leq 100$, joint observation transitions $T$: $\{((o_{1}, ..., o_{N}), (o'_{1}, ..., o'_{N}))_{1:sizeof(T)}\}$;
    \State \textbf{Output}: a set of multi-agent options $\Omega$;
    \State \textbf{Create} an empty set of options $\Omega$
    \State \textbf{Divide} the $N$ agents into sub-groups using Algorithm \ref{alg:Algo3}
    \For {every sub-group $G$}
        \State \textbf{Define} the agents in $G$ as $\{g_{1}, ..., g_{sizeof(G)}\}$
        \State \textbf{Extract} the set of transitions for $G$ from $T$ as $T_{G} \leftarrow $ 
        \Statex $\quad\ $ $\{(\underbrace{(o_{g_{1}}, ..., o_{g_{sizeof(G)}})}_{o_{G}}, \underbrace{(o'_{g_{1}}, ..., o'_{g_{sizeof(G)}})}_{o_{G}'})_{1:sizeof(T)}\}$
        \State \textbf{Learn} the connectivity function of $G$'s joint observa-
        \Statex $\quad$ tion space $f_{G}$ by minimizing $L(f_{G})$ (Equation (\ref{equ:1}))
        \State \textbf{Define} $\beta'$ as the $k$-th percentile value of $f_{G}$ in $T_{G}$
        \State \textbf{Acquire} the termination condition of option $\omega_{G}$: $$ \beta_{G}(o_{G}) \leftarrow \left\{
                \begin{aligned}
                1 &  & if \ f_{G}(o_{G}) < \beta' \\
                0 &  & otherwise
                \end{aligned}
                \right.
                $$  
        \State \textbf{Acquire} the initiation set of option $\omega_{G}$: $I_{G} \leftarrow$ 
        \Statex $\quad$ $\{o_{G}|\beta_{G}(o_{G})==0\}$
        \State \textbf{Learn} the policy $\pi_{G}$ of option $\omega_{G}$ by maximizing 
        \Statex $\quad$ the accumulated reward defined as Equation (\ref{equ:18}) using 
        \Statex $\quad$ MSAC based on $T_{G}$
        \State \textbf{Add} option $\omega_{G}$: $<I_{G}, \pi_{G}, \beta_{G}>$ to $\Omega$
    \EndFor
    \State \textbf{Return} $\Omega$
\end{algorithmic}}
\end{algorithm}

In order to take advantage of options in the learning process, we adopt a hierarchical RL framework shown as Algorithm \ref{alg:Algo1}. Typically, we train a RL agent to select among the primitive actions, aiming to maximize the accumulated reward. In our algorithm framework, we view this agent as a one-step option -- primitive option. As shown in Figure \ref{fig:0}, when getting a new observation, the hierarchical RL agent first decides on which option $\omega$ to use according to the high-level policy, and then decides on the action (primitive action) to take based on the corresponding intra-option policy $\pi_{\omega}$. In this hierarchical framework, there are two kinds of policy: the high-level policy for selecting among the options, and the low-level (intra-option) policy for selecting among the primitive actions. In this paper, we adopt multi-agent soft actor-critic (MSAC) \cite{DBLP:conf/icml/IqbalS19}, a typical MARL algorithm, to train the intra-option policy. However, the high-level policy is not executed step by step -- as mentioned in Section \ref{dcod}, once an option is taken, the actions will be selected according to $\pi_{\omega}$ until the option terminates, so we do not decide on a new option with the high-level policy until the previous option terminates. In this case, we propose \textbf{HA-MSAC}, a hierarchical MARL algorithm based on soft actor-critic, for training the high-level policy, which is introduced in Section \ref{HA-MSAC}.

Except for the primitive option, other options are extracted from transitions collected in the training process, so at first there is only the primitive option $A$ in the option set $\Omega'$ for selecting (Line 2 in Algorithm \ref{alg:Algo1}). Shown as Line 8-12 in Algorithm \ref{alg:Algo1}, every time when we get the set of new options $\Omega$ (at most for $N_{\omega}$ times), we merge it with $\{A\}$ to update $\Omega'$ (Line 10 in Algorithm \ref{alg:Algo1}). To construct the options, we extend Algorithm 1 of \cite{DBLP:conf/iclr/JinnaiPMK20} and Equation (\ref{equ:20}) from a single-agent scenario to a multi-agent scenario to obtain our Algorithm \ref{alg:Algo2} and Equation (\ref{equ:1}) (as described in Algorithm \ref{alg:Algo2}, $o_{G}$: the joint observation of sub-group $G$, $T_{G}$: the joint observation transitions of sub-group $G$). Through this extension, collaboration among the agents is considered not only at a high level, but also in the option discovery process, which can further improve the performance in scenarios where close collaboration is required.

\begin{equation} \label{equ:1}  
\begin{aligned}
    L(f_{G}) &= \frac{1}{2}\mathbb{E}_{(o_{G}, o_{G}')\sim T_{G}}[(f_{G}(o_{G})-f_{G}(o_{G}'))^{2}] + \eta\mathbb{E}_{o_{G}\sim\rho, o_{G}'\sim\rho}\\&[(f_{G}(o_{G})^{2}-1)(f_{G}(o_{G}')^{2}-1)+2f_{G}(o_{G})f_{G}(o_{G}')]
\end{aligned}
\end{equation}

As described in Algorithm \ref{alg:Algo2}, the reward function for learning the intra-option policy is based on $f_{G}$ which represents the connectivity of the joint observation space and is task-independent. Through this intrinsic reward term, agents are encouraged to visit the under-explored regions of their joint observation space. Further, inspired by \cite{DBLP:conf/atal/YangBZ20}, we can take advantage of the extrinsic reward that is specific to the environment to guarantee that the skills/options learned are also useful for team performance, so the reward at step $t$ for learning the intra-option policy can be defined as:
\begin{equation} \label{equ:18}
\begin{aligned}
    (r_{t}^{i})_{option} = (r_{t}^{i})_{env} +  \eta [f_{G_{i}}({o^{G_{i}}_{t}}) - f_{G_{i}}({o^{G_{i}}_{t+1}})]
\end{aligned}
\end{equation}
where $(r_{t}^{i})_{env}$ is task-related, $\eta$ is the weight for the intrinsic term, and $G_{i}$ is the sub-group that agent $i$ belongs to.

\subsection{Network Structure} \label{network}

In real-life multi-agent systems, the agents can usually be divided into some sub-groups and each sub-group is responsible for a sub-task (option). Therefore, before discovering multi-agent options, we first adopt the widely-used soft attention mechanism \cite{DBLP:conf/nips/VaswaniSPUJGKP17} to quantify the interaction among the agents from which we can abstract the sub-groups. Through this division, we can also avoid the joint observation space growing too large, so as to learn the multi-agent options more efficiently. Next, we introduce how to integrate the attention mechanism in the network design.

The network structure of the high/low-level policy is shown in Figure \ref{fig:1(a)}. When deciding on which multi-agent option is available or to execute, agent $i$ needs to know other agents' observations (joint observation) to see whether they have reached the initiation set, so the communication among the agents is required. Further, we view this multi-agent system as a fully-connected directional graph, where each node represents an agent and the weight of edge $i \to j$ represents the importance of agent $j$ to agent $i$. These weights can be learnt through a soft attention mechanism (i.e., a query-key system). After that, we adopt GNN (e.g., weighted summation) to extract $x_{i}$, the contribution from other agents to agent $i$, and then concatenate it with its own observation embedding $h_{i}$ as the input of its own policy head $\pi_{i}$. In this way, $\pi_{i}$ is based on observations of all the agents, and the attention mechanism can filter out observations from the agents that have no/low interaction with agent $i$ for better decision-making. 

The Q-function network of the high/low-level policy is shown in Figure \ref{fig:1(b)}. We only use the observations $o_{1:N}$ to extract the interaction relationship rather than the observation-option/action pairs $(o, \omega)_{1:N}$ (like in \cite{DBLP:conf/aaai/LiuWHHC020, DBLP:conf/icml/IqbalS19}), because: (1) the interaction relationship extracted by the policy and Q-function network should be the same, since it's for the same multi-agent task with the same group of agents; (2) parameters of the attention mechanism can be shared between the two networks, which can improve the training efficiency. Also, note that the parameters of the soft attention and GNN part: $W_{k}, W_{q}, W_{v}, W^{\omega}_{v}$ are shared by all the agents to encourage a common embedding space.

The attention module in the intra-option policy of the primitive option (defined in Section \ref{framework}) is used for the sub-group division. As shown in Algorithm \ref{alg:Algo3}, we use the transitions collected in the training process as input and calculate the accumulated soft attention weight matrix based on the output of the attention layer. Then, we compare the normalized attention weights with a predefined threshold $z$ to finalize the sub-group for each agent. In our experiment, the threshold $z$ is set as $1/(N-1)$, where $N$ is the number of the agents in the learning system. The intuition is that the attention for the agents within the same sub-group should be higher than the average level. The threshold can also be fine-tuned as a hyperparameter. For example, a higher threshold $z$ may be required for the more closely collaborative tasks.

\begin{algorithm}[t]
\caption{Sub-group Division}\label{alg:Algo3}
{\begin{algorithmic}[1]
    \State \textbf{Input}: threshold $z \in (0, 1)$, a set of joint observations $T$: $\{(o_{1}, ..., o_{N})_{1:sizeof(T)}\}$, policy network $\pi$;
    \State \textbf{Output}: a set of sub-groups;
    \For{$t=1$ to $sizeof(T)$ }
        \State \textbf{Feed} $(o_{1}, ..., o_{N})_{t}$ into $\pi$
        \State \textbf{Save} the soft attention weights $(W^{i,j}_{S})_{t}$, where $i \in $
        \Statex $\quad$ $[1, N]$, $j \in [1, N]$ and $j \neq i$
    \EndFor
    \For{each agent $i$}
        \State \textbf{Insert} $i$ to its sub-group $G_{i}$
        \For{each agent $j \neq i$}
            \If{$\sum_{t}(W_{S}^{i,j})_{t} \geq z*\sum_{n \neq i}\sum_{t}(W_{S}^{i,n})_{t}$}
                \State \textbf{Insert} $j$ to $G_{i}$
            \EndIf
        \EndFor
    \EndFor
    \State \textbf{Collect} $G \leftarrow \{G_{1},...,G_{N}\}$ and eliminate duplicate elements
    \State \textbf{Return} $G$ 
\end{algorithmic}}
\end{algorithm}

\section{Hierarchical Attentive Multi-agent Soft Actor-Critic (HA-MSAC)} \label{HA-MSAC}

\begin{figure*}[!htbp]

	\centering
	\includegraphics[width=7.0in, height=1.3in]{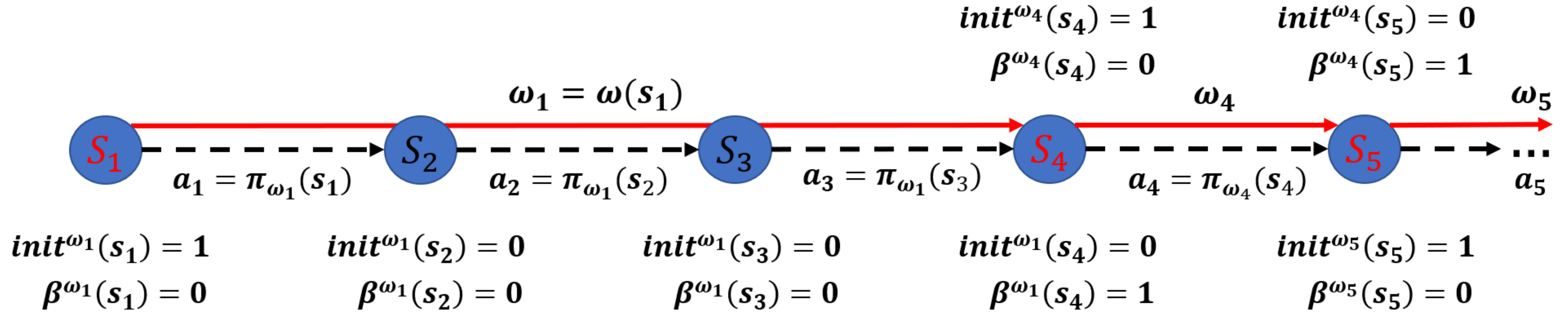}
	\caption{Single-agent Scenario: The agent chooses the option $\omega_{1}$ at $s_{1}$ ($init^{\omega_{1}}(s_{1})==1$), then the corresponding intra-option policy $\pi_{\omega_{1}}$ is executed until $s_{4}$ ($\beta^{\omega_{1}}(s_{4})==1$). Meanwhile, a new option $\omega_{4}$ is chosen at $s_{4}$ ($init^{\omega_{4}}(s_{4})==1$). Note that the high-level policy is not executed at state $s$ until the previous option $\omega$ ends at this step ($\beta^{\omega}(s)==1$) and a new option is required.}
	\label{fig:3}

\end{figure*}

\begin{figure*}[!htbp]

	\centering
	\includegraphics[width=7.0in, height=1.9in]{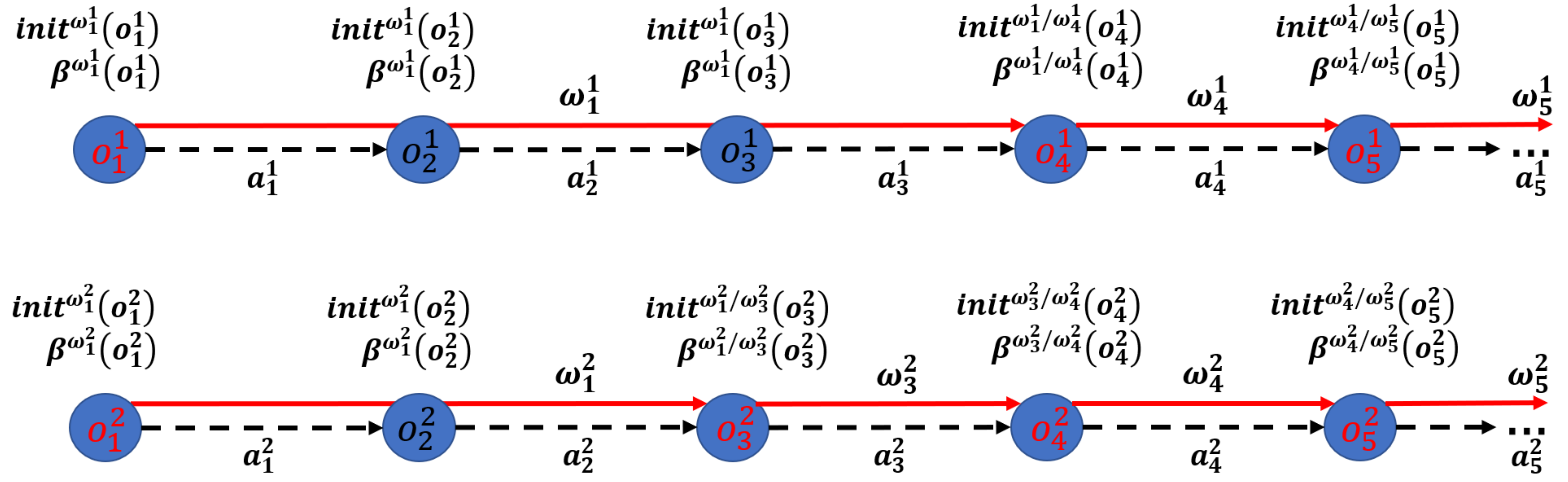}
	\caption{Multi-agent Scenario: Similar with the single-agent scenario, $init^{\omega^{i}_{t}}(o^{i}_{t})==1$, if agent $i$ starts to execute the option $\omega^{i}_{t}$ when getting the observation $o^{i}_{t}$ at step $t$; $\beta^{\omega^{i}_{t}}(o^{i}_{t'})==1$, if the option $\omega^{i}_{t}$ ends at step $t'$. However, the option choice among the agents maybe asynchronous, which brings new difficulty as compared with the single-agent scenario. For example, agent 2 starts a new option at step 3, while agent 1 is still within $\omega^{1}_{1}$, so the two agents' high-level policy are not executed at the same time. Note that when updating the high-level policy of agent $i$, the gradient calculated based on the transition at step $t$ can be applied only if $init^{\omega^{i}_{t}}(o^{i}_{t})==1$, which means the high-level policy of agent $i$ is executed at this step.}
	\label{fig:4}
\end{figure*}

In this section, we extend MSAC \cite{DBLP:conf/icml/IqbalS19}, a widely-used MARL algorithm with \textbf{attentive critics}, to a hierarchical multi-agent reinforcement learning algorithm with \textbf{attentive actors and critics}, which we name as \textbf{HA-MSAC}. This algorithm can be used to update the high-level policy and corresponding Q-function network in an off-policy manner.

\subsection{Extension with Attention Modules}

In MSAC, they extended soft actor-critic (SAC) \cite{DBLP:conf/icml/HaarnojaZAL18} to a multi-agent scenario with \textbf{attentive critics}, and define the objective functions for training the policy and Q-function network of agent $i$ as follows: ($o = (o_{1}, ..., o_{N}), a = (a_{1}, ..., a_{N})$)

\begin{equation} \label{equ:2}
\begin{aligned}
    L_{Q}(\phi)=\sum_{i=1}^{N}\mathbb{E}_{(o,a,r,o')\sim B}[(Q_{i}^{\phi}(o,a)-y_{i})^{2}]
\end{aligned}
\end{equation}

\begin{equation} \label{equ:3}
\begin{aligned}
    y_{i}=r_{i}+\gamma\mathbb{E}_{a'\sim\pi_{\overline{\theta}}(o')}[Q_{i}^{\overline{\phi}}(o',a')-\alpha log(\pi_{\overline{\theta}_{i}}(a_{i}'|o_{i}'))]
\end{aligned}
\end{equation}

\begin{equation} \label{equ:4}
\begin{aligned}
    \nabla_{\theta_{i}}J(\pi_{\theta_{i}})=\mathbb{E}_{o\sim B, a\sim \pi_{\theta}(o)}[\nabla_{\theta_{i}}log(\pi_{\theta_{i}}(a_{i}|o_{i}))\\
    (-\alpha log(\pi_{\theta_{i}}(a_{i}|o_{i}))+Q_{i}^{\phi}(o,a)-b(o,a_{\setminus i}))]
\end{aligned}
\end{equation}

\begin{equation} \label{equ:5}
\begin{aligned}
    b(o,a_{\setminus i}) &= \mathbb{E}_{a_{i}\sim \pi_{\theta_{i}}(o_{i})}[Q_{i}^{\phi}(o,(a_{i}, a_{\setminus i}))] \\
    &= \sum_{a_{i} \in A_{i}}\pi_{\theta_{i}}(a_{i}| o_{i})Q_{i}^{\phi}(o,(a_{i}, a_{\setminus i}))
\end{aligned}
\end{equation}
where $B$ represents the replay buffer, $\overline{\theta}$ and $\overline{\phi}$ are the parameters of the target policy and Q-function network, $\alpha log(\pi_{\theta})$ is the entropy term. $b$ is the baseline term of the advantage function, and in the case of discrete policies, it can be calculated explicitly (Equation (\ref{equ:5})). However, we have to estimate the expectation in Equation (\ref{equ:3}) and (\ref{equ:4}) through sampling. For example, in Equation (\ref{equ:3}), we can't enumerate all the possible joint actions $a'$, so we calculate $a_{i}' \sim \pi_{\overline{\theta}_{i}}(o_{i}'),\ for\ i=1\ to\ N$, and then obtain the sampled joint action $a'$.

In our setting, besides the attentive critics, we also use attentive actors (shown as Figure \ref{fig:1(a)}), which means that each agent's policy takes the observations of all the agents as input, so we modify the calculation of $y_{i}$, $\nabla_{\theta}J(\pi_{\theta})$, and $b(o,a_{\setminus i})$ as follows:

\begin{equation} \label{equ:6}
\begin{aligned}
    y_{i}=r_{i}+\gamma\mathbb{E}_{a'\sim\pi_{\overline{\theta}}(o')}[Q_{i}^{\overline{\phi}}(o',a')-\alpha log(\pi_{\overline{\theta}_{i}}(a_{i}'|o'))]
\end{aligned}
\end{equation}

\begin{equation} \label{equ:7}
\begin{aligned}
    \nabla_{\theta}J(\pi_{\theta})=\sum_{i=1}^{N}\mathbb{E}_{o\sim B, a\sim \pi_{\theta}(o)}[\nabla_{\theta_{i}}log(\pi_{\theta_{i}}(a_{i}|o))\\
    (-\alpha log(\pi_{\theta_{i}}(a_{i}|o))+Q_{i}^{\phi}(o,a)-b(o,a_{\setminus i}))]
\end{aligned}
\end{equation}

\begin{equation} \label{equ:8}
\begin{aligned}
    b(o,a_{\setminus i}) &= \mathbb{E}_{a_{i}\sim \pi_{\theta_{i}}(o)}[Q_{i}^{\phi}(o,(a_{i}, a_{\setminus i}))] \\
    &= \sum_{a_{i} \in A_{i}}\pi_{\theta_{i}}(a_{i}| o)Q_{i}^{\phi}(o,(a_{i}, a_{\setminus i}))
\end{aligned}
\end{equation}
Note that when estimating the expectation in Equation (\ref{equ:6}) and (\ref{equ:7}), we don't need any extra sampling process as compared with Equation (\ref{equ:3}) and (\ref{equ:4}).

\subsection{Extension with Hierarchical MARL}

As mentioned in Section \ref{framework}, the high-level policy is adopted to select among the options. Once an option is chosen by the high-level policy, the corresponding intra-option policy will be executed until its termination condition is satisfied. For example, in Figure \ref{fig:3}, the agent chooses the option $\omega_{1}$ at $s_{1}$ ($init^{\omega_{1}}(s_{1})==1$), then the intra-option policy $\pi_{\omega_{1}}$
is executed until $s_{4}$ ($\beta^{\omega_{1}}(s_{4})==1$). Therefore, the execution of the high-level policy is not step-by-step but option-by-option and each option can last for steps, which is different from typical RL defined on the MDP. Next, we describe the objective functions for learning with options. As a reminder, we first transfer the option-critic framework \cite{DBLP:conf/atal/ChakravortyWRCB20} to SAC, and then extend it from the single-agent scenario to the multi-agent scenario. 

Firstly, as shown in Figure \ref{fig:3}, during the sampling process,we can collect transitions like $(s_{t},\ \omega_{t},\ init^{\omega_{t}}_{t},\ a_{t},\ r_{t},\ s_{t+1},\ \beta^{\omega_{t}}_{t+1})$, where $\omega_{t}$ is the option choice at step $t$, $init^{\omega_{t}}$ and $\beta^{\omega_{t}}$ are the corresponding initiation and termination signal. Based on them, we can define the objective functions for the single-agent scenario as follows:
\begin{equation} \label{equ:9}
\begin{aligned}
    L_{Q}(\phi)=\mathbb{E}_{(s_{t},\omega_{t},r_{t}, s_{t+1},\beta^{\omega_{t}}_{t+1})\sim B}[(Q^{\phi}(s_{t},\omega_{t})-y_{t})^{2}]
\end{aligned}
\end{equation}
\begin{equation} \label{equ:10}
\begin{aligned}
    &y_{t}=r_{t}+\gamma[(1-\beta^{\omega_{t}}_{t+1})Q^{\overline{\phi}}(s_{t+1},\omega_{t})+\\
    &\beta^{\omega_{t}}_{t+1}\mathbb{E}_{\omega_{t+1}\sim\pi_{\overline{\theta}}(s_{t+1})}(Q^{\overline{\phi}}(s_{t+1},\omega_{t+1}) - \alpha log(\pi_{\overline{\theta}}(\omega_{t+1}|s_{t+1})))]
\end{aligned}
\end{equation}
Intuitively, if $\omega_{t}$ terminates at $s_{t+1}$ ($\beta^{\omega_{t}}_{t+1}==1$), the agent will choose a new option based on the target high-level policy $\pi_{\overline{\theta}}$; otherwise, it will still execute $\omega_{t}$ (i.e., $\omega_{t+1}==\omega_{t}$). 

While, the objective function for updating the policy network is given as:
\begin{equation} \label{equ:11}
\begin{aligned}
    \nabla_{\theta}J(\pi_{\theta})=&\mathbb{E}_{(s_{t}, init_{t})\sim B, \omega_{t}\sim \pi_{\theta}(s_{t})}[init_{t} \ast \nabla_{\theta}log(\pi_{\theta}(\omega_{t}|s_{t}))
    \\
    &(-\alpha log(\pi_{\theta}(\omega_{t}|s_{t}))+Q^{\phi}(s_{t},\omega_{t})-b(s_{t}))]
\end{aligned}
\end{equation}
The intuition behind this equation is that $init_{t}$ marks the steps where the high-level policy $\pi_{\theta}$ is executed ($init_{t}==1$) and thus the computed gradient is effective for update. 

Secondly, as shown in Figure \ref{fig:4}, when extended to the multi-agent scenario (with $N$ agents), we can collect the transitions like: $(o_{t},\ \omega_{t}, \ init_{t},\ a_{t},\ r_{t},\ o_{t+1},\ \beta_{t+1}) = [(o^{1}_{t},...,o^{N}_{t}),\ (\omega^{1}_{t},...,\omega^{N}_{t}),\ (init^{\omega^{1}_{t}}_{t},...,init^{\omega^{N}_{t}}_{t}),\ \\(a^{1}_{t},...,a^{N}_{t}),\ (r^{1}_{t},...,r^{N}_{t}),\ (o^{1}_{t+1},...,o^{N}_{t+1}),\ (\beta^{\omega^{1}_{t}}_{t+1},...,\beta^{\omega^{N}_{t}}_{t+1})]$. Note that the option choice among the agents maybe asynchronous and this brings new difficulty. For example, in Figure \ref{fig:4}, agent 2 starts a new option at step 3, while agent 1 is still within $\omega^{1}_{1}$, so we cannot update the two agents' high-level policy at the same time. Instead, the update should be based on the termination or initiation signal specific to each agent. Thus, we define its objective functions as follows:
\begin{equation} \label{equ:12}
\begin{aligned}
    L_{Q}(\phi)=\sum_{i=1}^{N}\mathbb{E}_{(o_{t},\omega_{t}, r_{t},o_{t+1},\beta_{t+1})\sim B}[(Q_{i}^{\phi}(o_{t},\omega_{t})-y_{t}^{i})^{2}]
\end{aligned}
\end{equation}
\begin{equation} \label{equ:13}
\begin{aligned}
    y_{t}^{i}=r_{t}^{i}+\gamma\mathbb{E}_{\omega_{t+1}\sim\pi_{\overline{\theta}}(o_{t+1})}[Q_{i}^{\overline{\phi}}(o_{t+1},\omega_{t+1})\\-\alpha \beta^{\omega^{i}_{t}}_{t+1}log(\pi_{\overline{\theta}_{i}}(\omega_{t+1}^{i}|o_{t+1}))]
\end{aligned}
\end{equation}
When estimating the expectation in Equation (\ref{equ:13}), we need to sample the joint option $\omega_{t+1}$, which is generated in this way -- for $i=1$ to $N$: if $\beta_{t+1}^{\omega_{t}^{i}}==1$, $\omega_{t+1}^{i}\sim \pi_{\overline{\theta}_{i}}(o_{t+1})$; otherwise, $\omega_{t+1}^{i}=\omega_{t}^{i}$. The intuition is that, for agent $i$, the new option is not required to be sampled until its previous option terminates ($\beta_{t+1}^{\omega_{t}^{i}}==1$). 

While, the objective function for updating the policy network is given as:
\begin{equation} \label{equ:14}
\begin{aligned}
    \nabla_{\theta}J(\pi_{\theta})=\sum_{i=1}^{N}\mathbb{E}_{(o_{t}, \omega_{t},init_{t})\sim B, \hat{\omega}_{t}\sim \pi_{\theta}(o_{t})}[init_{t}^{i}\ast\\\nabla_{\theta_{i}}log(\pi_{\theta_{i}}(\hat{\omega}_{t}^{i}|o_{t}))
    (-\alpha log(\pi_{\theta_{i}}(\hat{\omega}_{t}^{i}|o_{t}))+\\Q_{i}^{\phi}(o_{t},\hat{\omega}_{t})-b(o_{t},\hat{\omega}_{t}^{\setminus i}))]
\end{aligned}
\end{equation}
\begin{equation} \label{equ:15}
\begin{aligned}
    b(o_{t},\hat{\omega}_{t}^{\setminus i}) = \mathbb{E}_{\hat{\omega}_{t}^{i}\sim \pi_{\theta_{i}}(o_{t})}[Q_{i}^{\phi}(o_{t},(\hat{\omega}^{i}_{t}, \hat{\omega}^{\setminus i}_{t}))]
\end{aligned}
\end{equation}
where the expectation in Equation (\ref{equ:15}) is calculated in the same way as Equation (\ref{equ:8}), while the expectation in Equation (\ref{equ:14}) is estimated by sampling $\hat{\omega}_{t}$ -- for $i=1$ to $N$: if $init_{t}^{i}==1$, $\hat{\omega}_{t}^{i}\sim \pi_{\theta_{i}}(o_{t})$; otherwise, $\hat{\omega}_{t}^{i}=\omega_{t}^{i}$. The intuition behind this that we need to use $init_{t}^{i}$ to mask out the steps where the high-level policy $\pi_{\theta_{i}}$ is not executed ($init_{t}^{i}==0$). To sum up, the objective functions of \textbf{HA-MSAC} are listed as Equation (\ref{equ:12})-(\ref{equ:15}).

\begin{figure}[t]
	\centering
    \includegraphics[height=2.3in, width=3.0in]{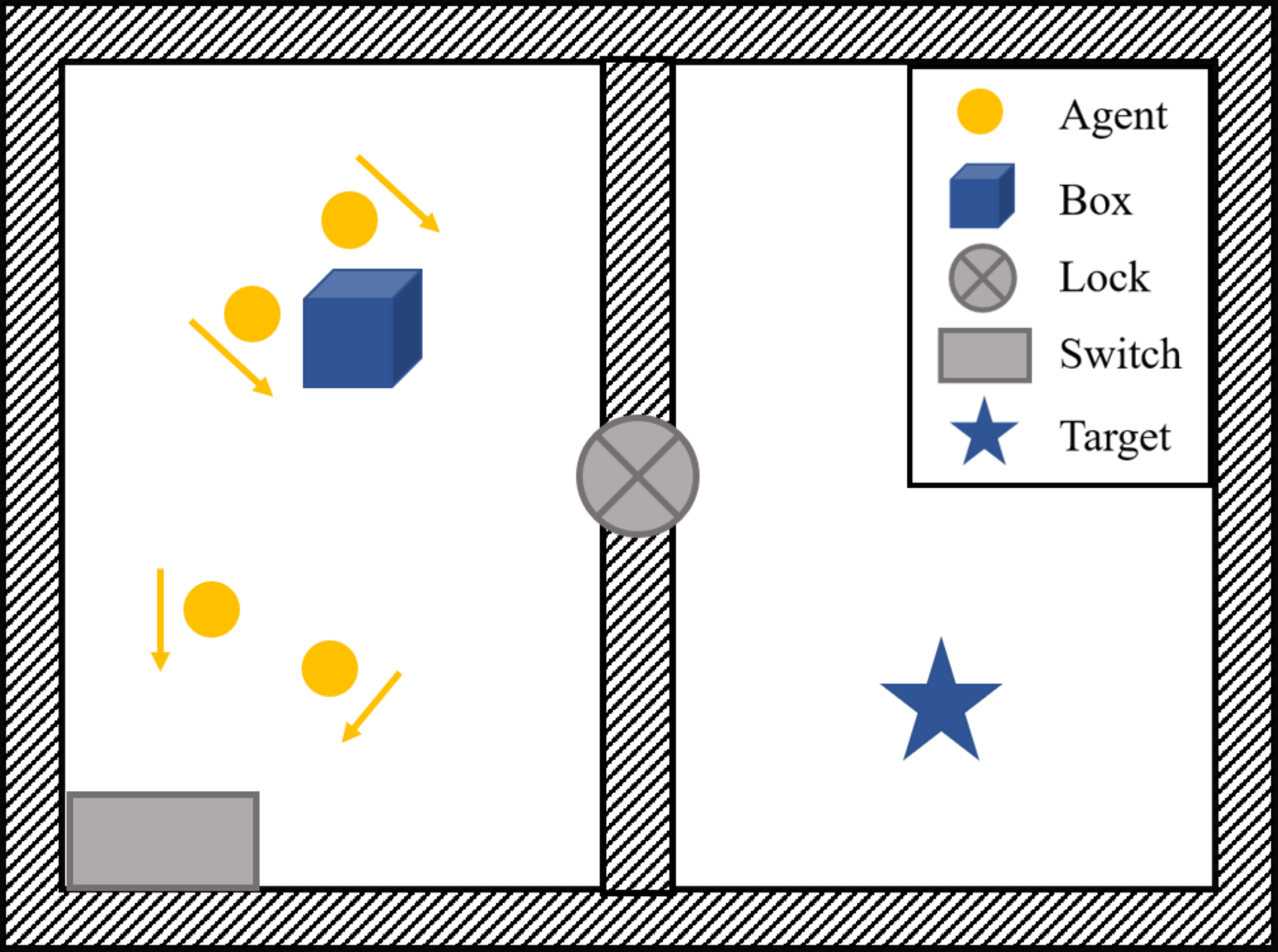}
	\caption{Task Scenario for Evaluation: There are in total four ``agents". Two (or more) ``agents" need to touch the ``switch" at the same time to turn on the ``switch" first, and then two (or more) ``agents" need to push the ``box" in the same direction at the same time to move it through the ``lock" and reach the ``target" location.}
	\label{fig:5}
\end{figure}

\begin{figure*}[t]
\centering
\subfigure[Training reward of agent 0]{
\label{fig:6(a)} 
\includegraphics[width=2.80in, height=1.70in]{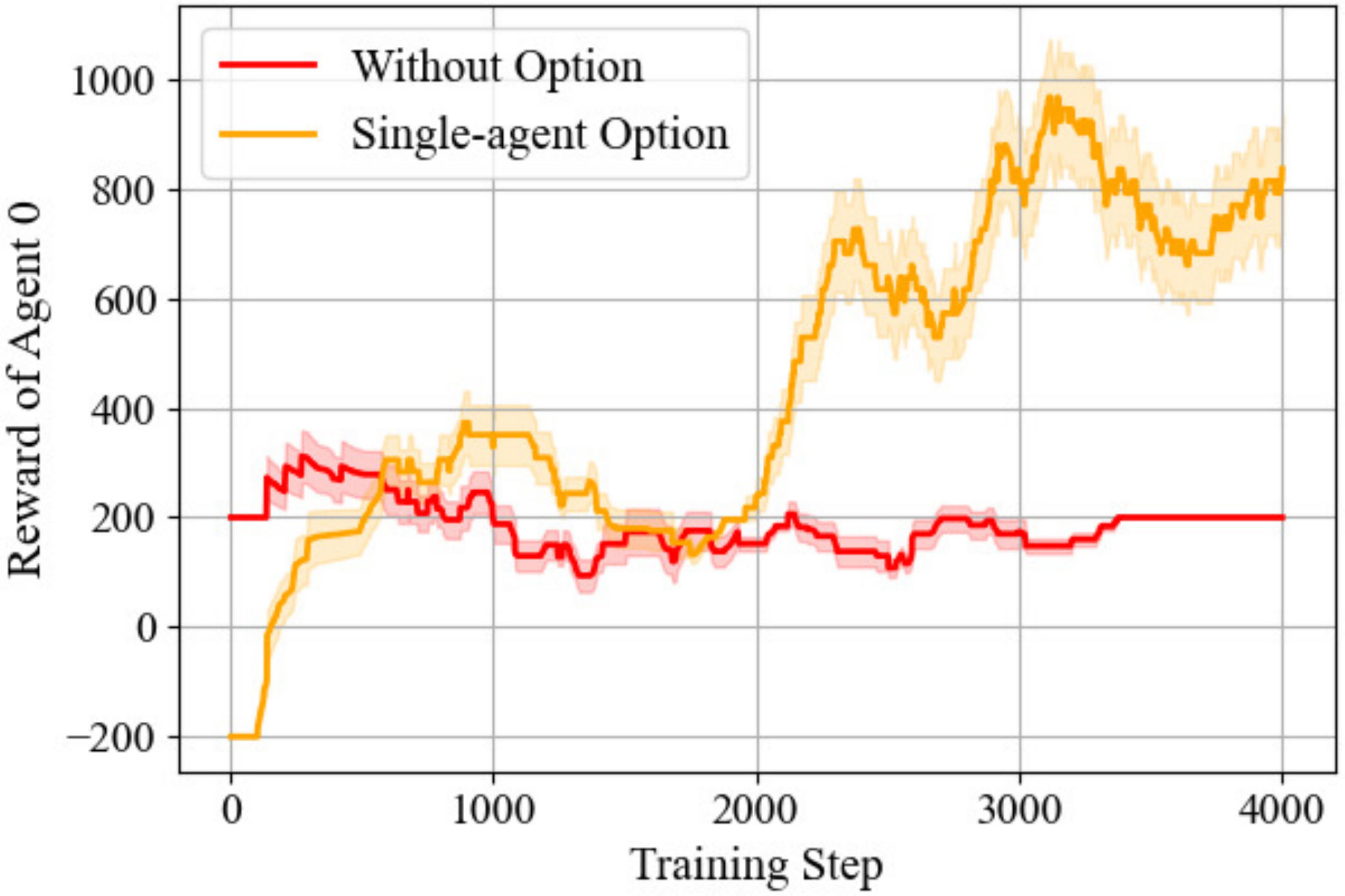}}
\subfigure[Training reward of agent 1]{
\label{fig:6(b)} 
\includegraphics[width=2.80in, height=1.70in]{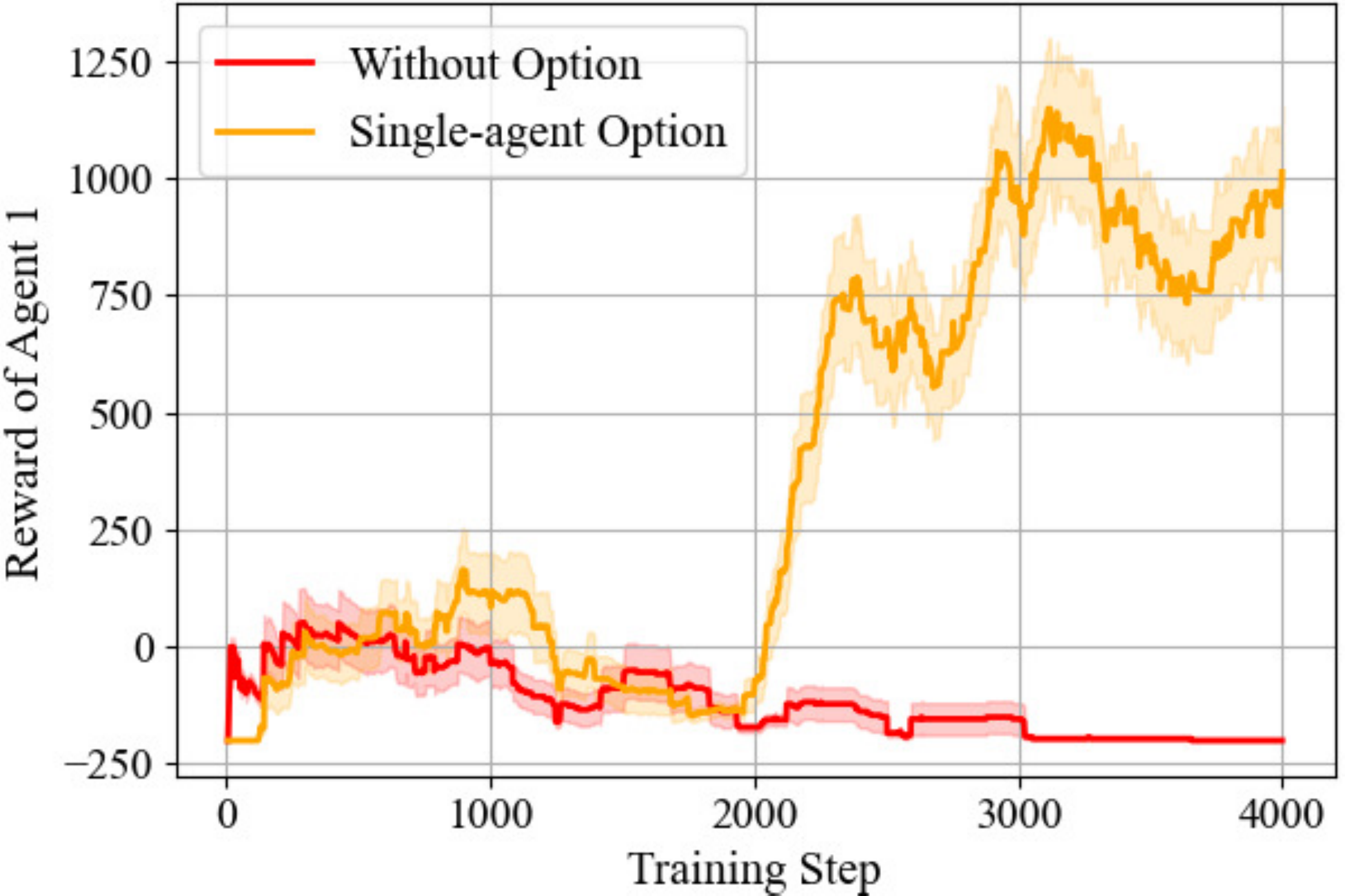}}

\caption{Sub-setting 1: There are only two agents -- agent 0 needs to turn on the switch first, then agent 1 can push the box to the target. The agent without options can learn how to turn on the switch and get the corresponding reward term, however, it cannot learn the harder task -- pushing the box to the target. After training a single-agent option for each agent at step 2000, the performance starts to improve, showing the effectiveness of MARL with single-agent options.
}

\label{fig:6} 
\end{figure*}

\begin{figure*}[t]
\centering
\subfigure[Training reward of agent 0 / 1]{
\label{fig:6(c)} 
\includegraphics[width=2.80in, height=1.70in]{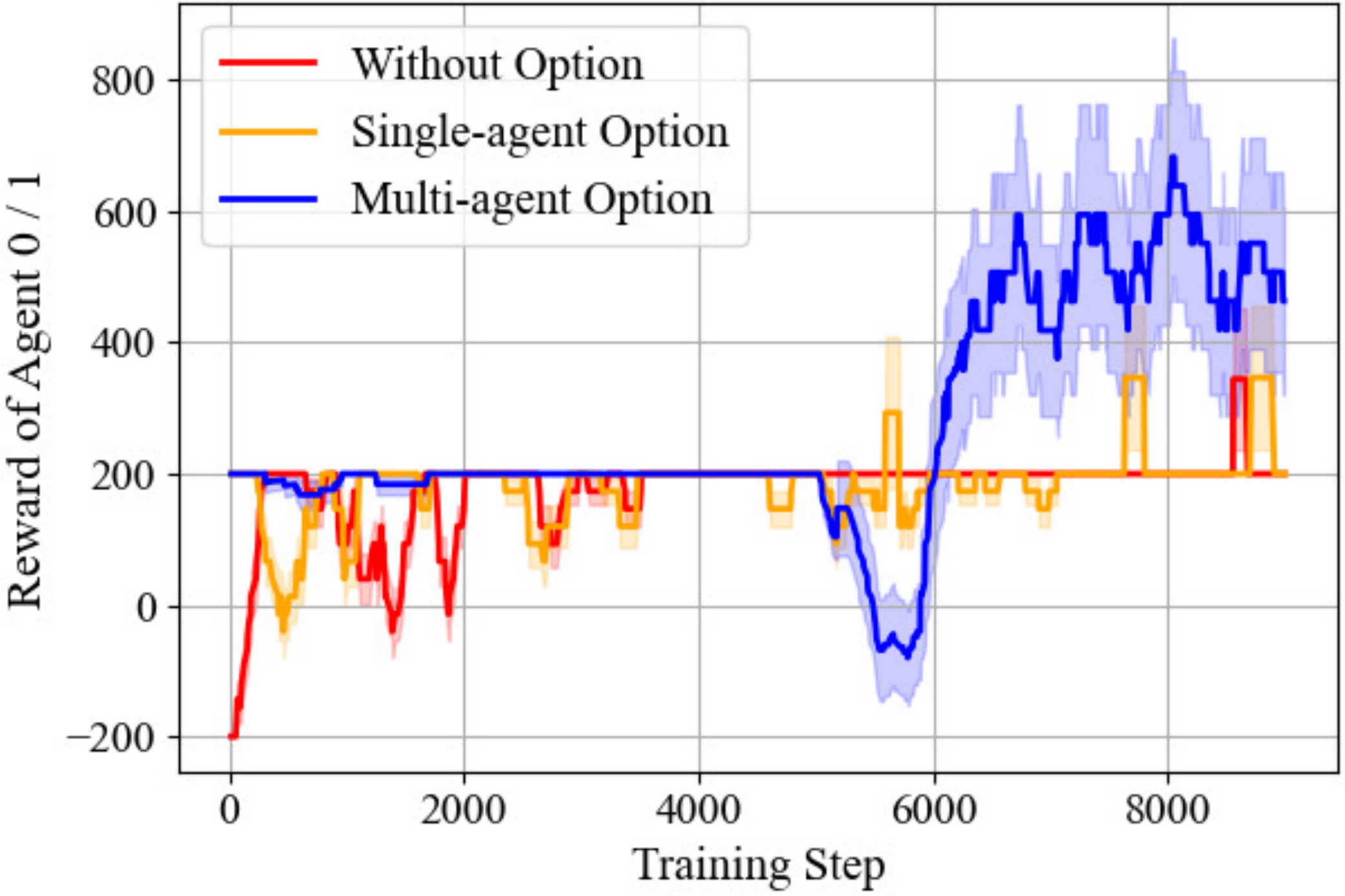}}
\subfigure[Training reward of agent 2 / 3]{
\label{fig:6(d)} 
\includegraphics[width=2.80in, height=1.70in]{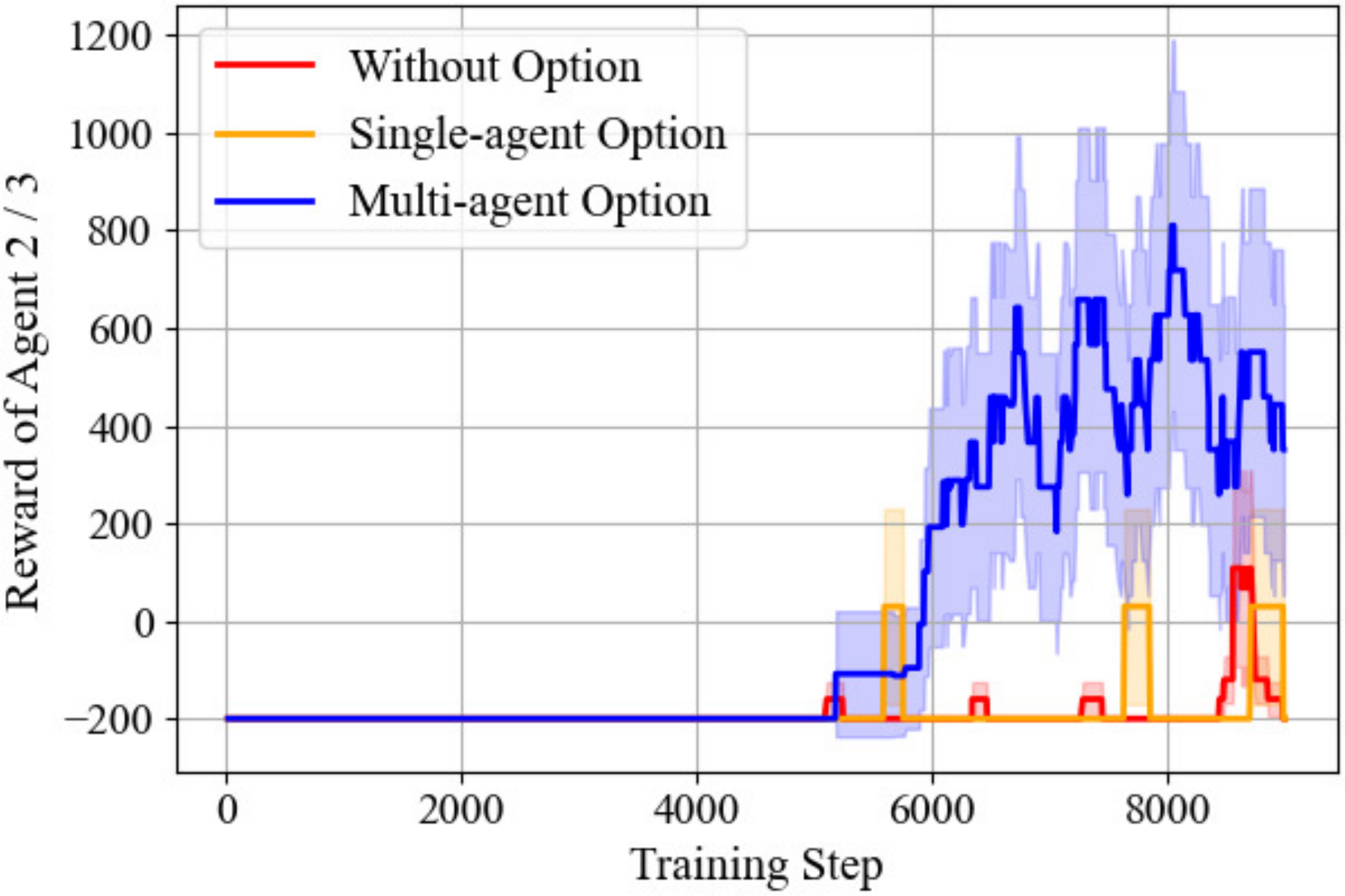}}

\caption{Sub-setting 2: There are four agents -- agent 0 and 1 need to collaborate with each other to turn on the switch first, then agent 2 and 3 need to work together to push the box to the target. Since collaboration is required for each sub-task, the agents without options or with single-agent options cannot complete the whole task very well, especially for pushing the box, as compared with the agents with multi-agent options. Note that there are two sub-groups (i.e., [0, 1] and [2, 3]), and a multi-agent option is trained for each sub-group to complete its sub-task at step 5000.}

\label{fig:8} 
\end{figure*}

\begin{figure*}[t]
\centering
\subfigure[Attention weights of sub-setting 2]{
\label{fig:6(e)} 
\includegraphics[width=2.80in, height=1.70in]{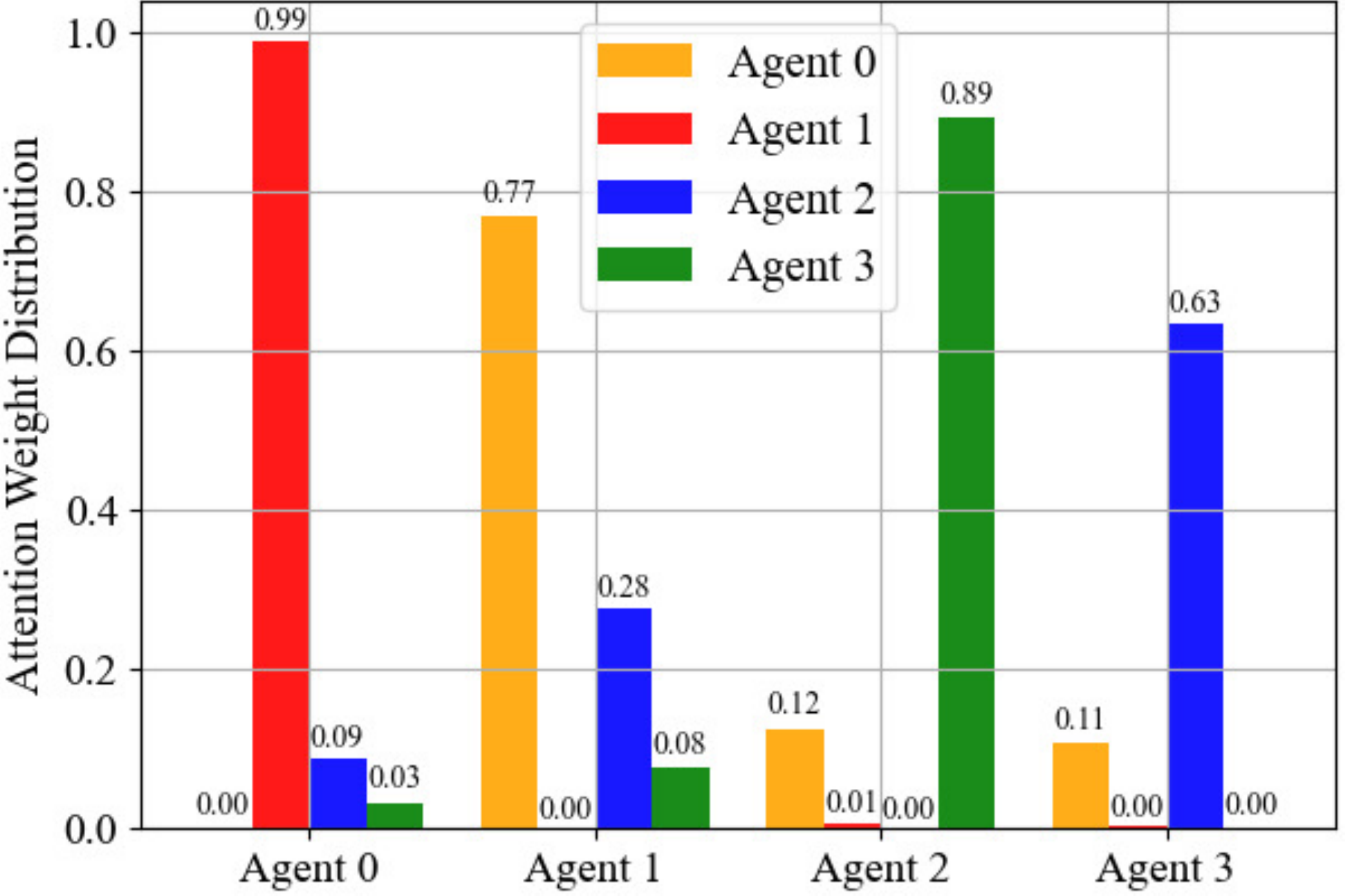}}
\subfigure[Attention weights of sub-setting 3]{
\label{fig:6(f)} 
\includegraphics[width=2.80in, height=1.70in]{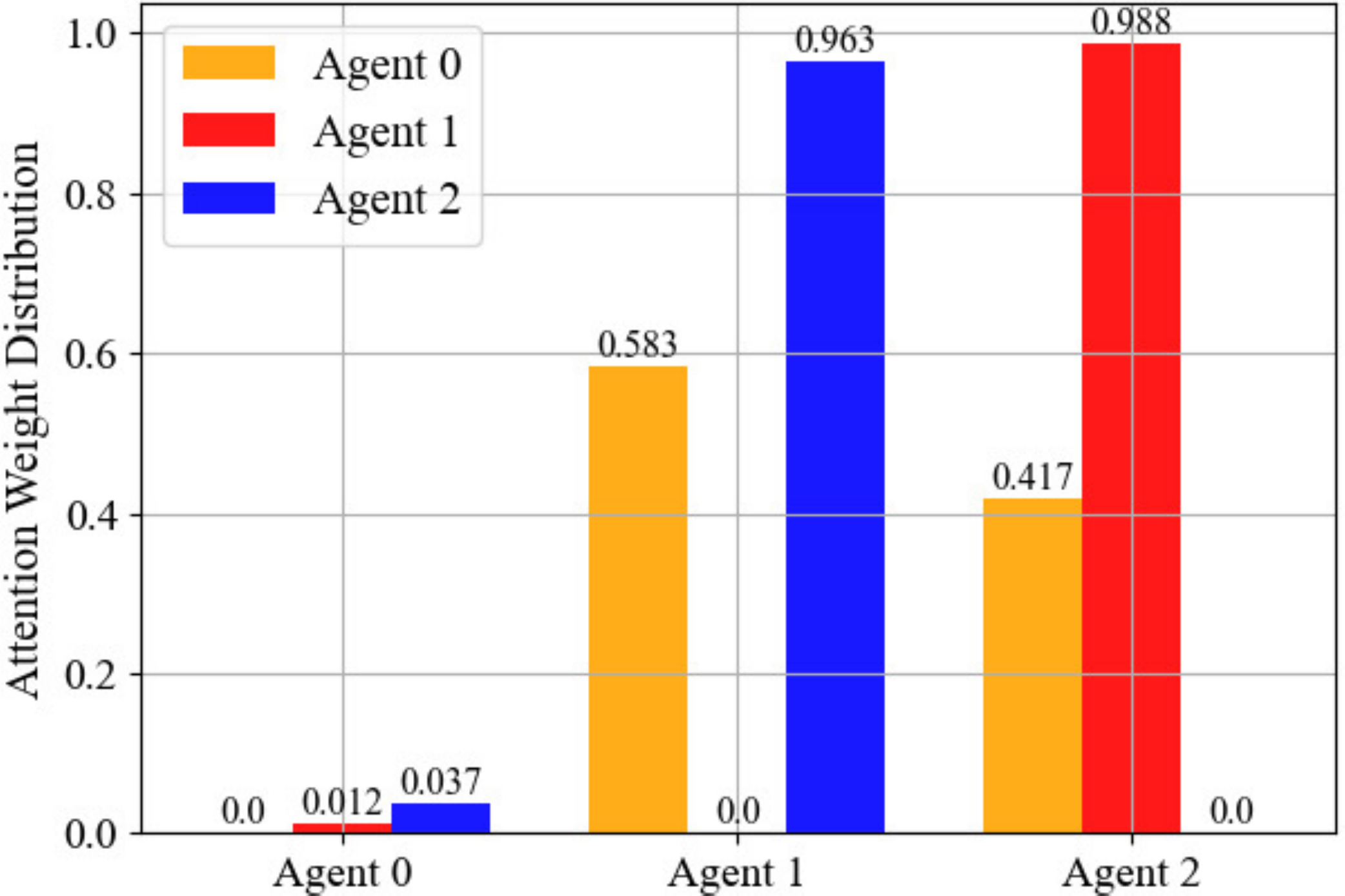}}

\caption{Attention weight distribution of sub-setting 2 and 3. For agent $i$, if the normalized attention weight of agent $j$ is higher than the threshold $z$, agent $i$ should form a sub-group with agent $j$. $z$ is set as $1/(N-1)$, where $N$ is the number of agents in the learning system. Therefore, we can get the sub-group divisions -- sub-setting 2: [0, 1] and [2, 3], sub-setting 3: [0, 1] and [1, 2], according to Algorithm \ref{alg:Algo3}. Note that these sub-group divisions are reasonable, since they are in line with the sub-task divisions of the corresponding sub-settings.}

\label{fig:9} 
\end{figure*}

\begin{figure*}[t]
\centering
\subfigure[Training reward of agent 0]{
\label{fig:7(a)} 
\includegraphics[width=2.30in, height=1.50in]{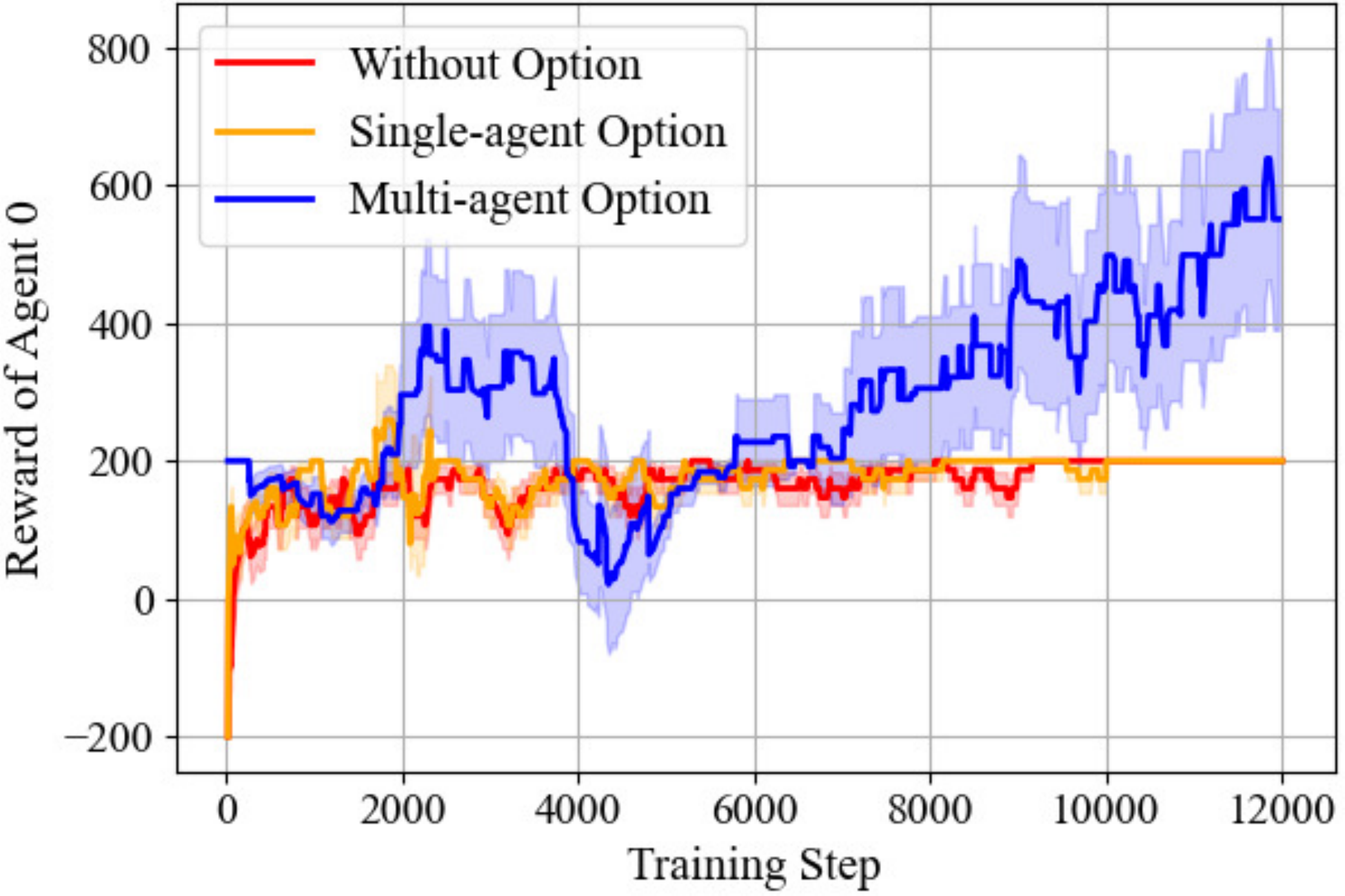}}
\subfigure[Training reward of agent 1]{
\label{fig:7(b)} 
\includegraphics[width=2.30in, height=1.50in]{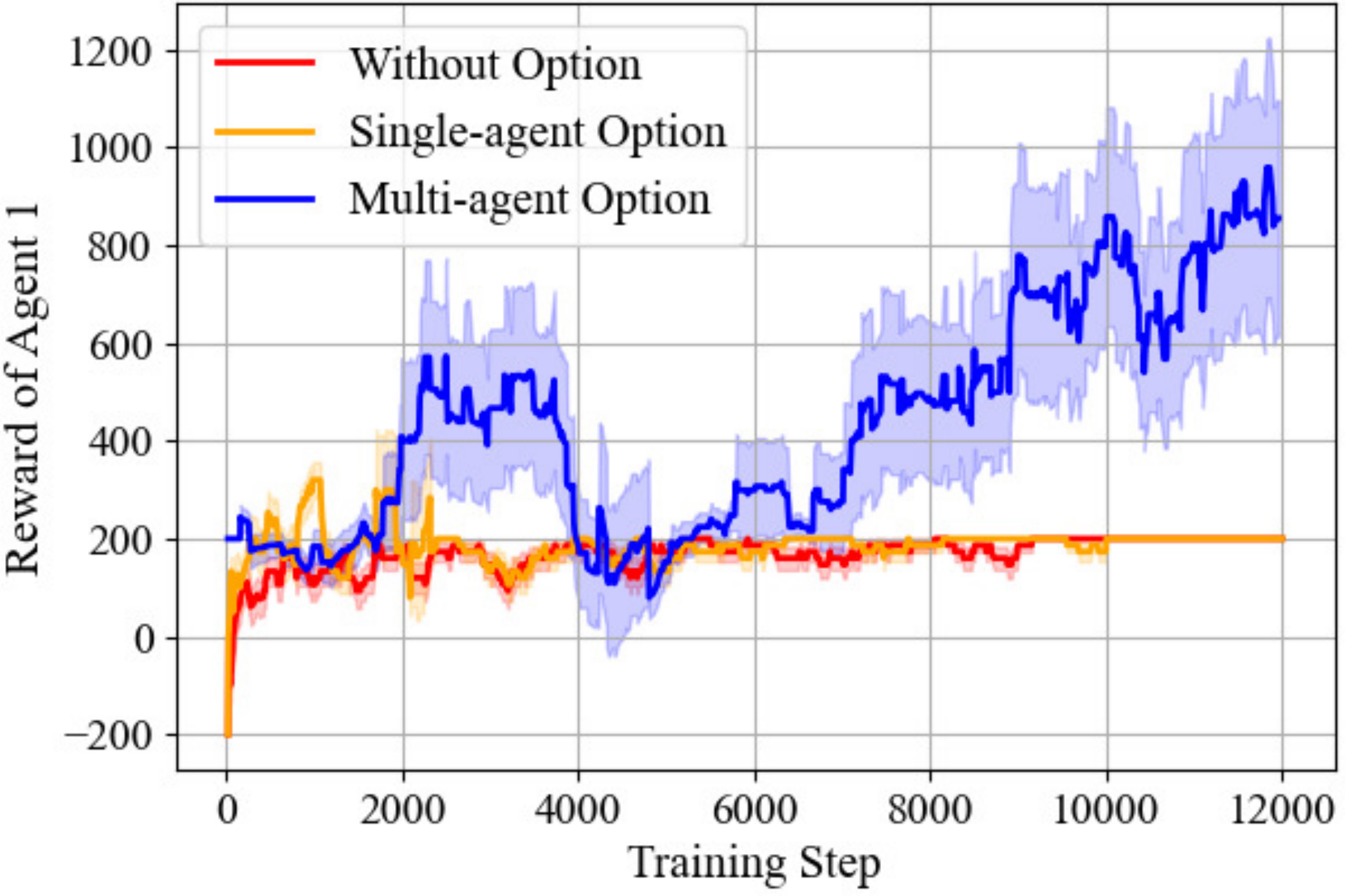}}
\subfigure[Training reward of agent 2]{
\label{fig:7(c)} 
\includegraphics[width=2.30in, height=1.50in]{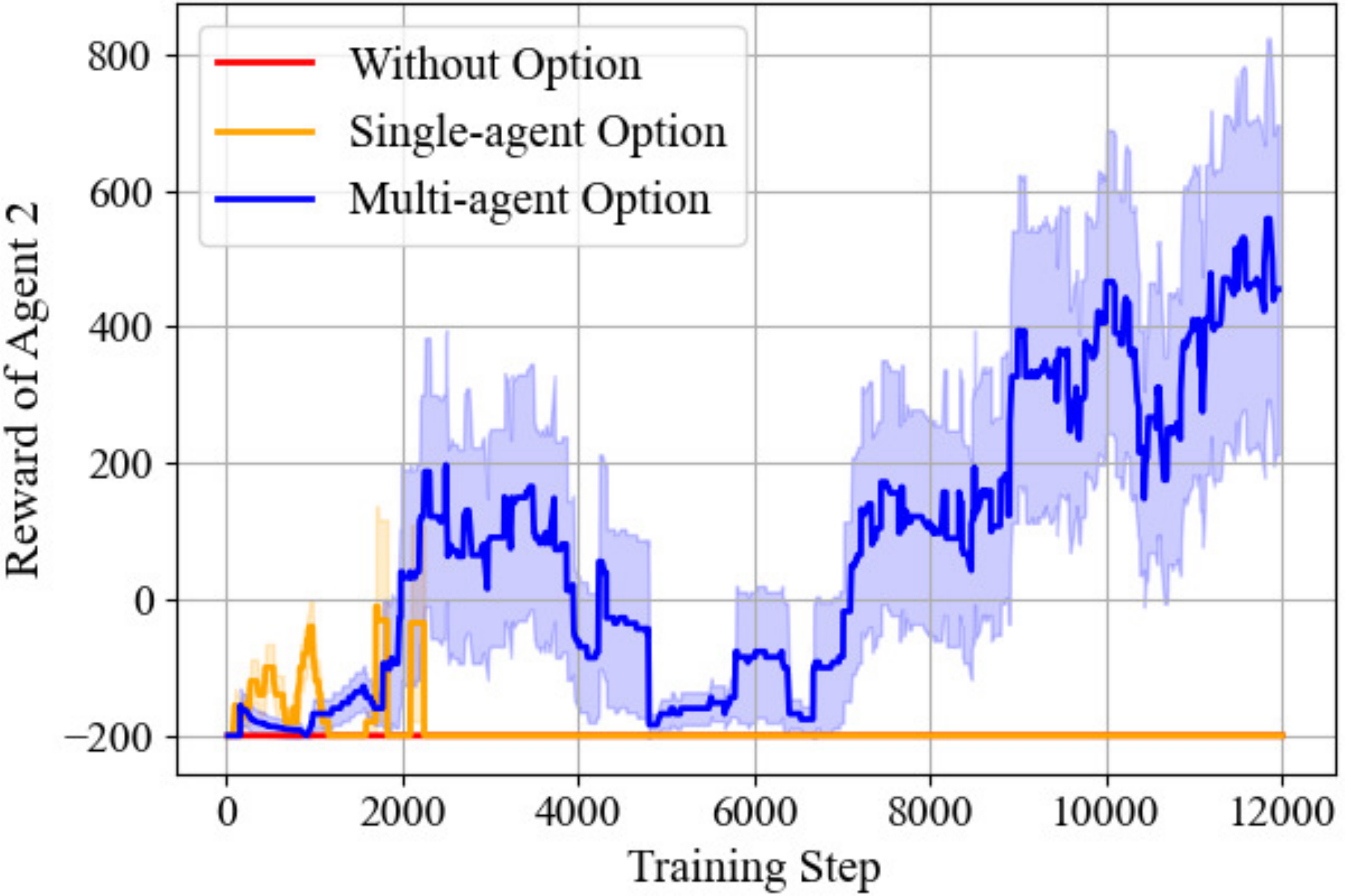}}

\caption{Sub-setting 3: There are three agents -- agent 1 should cooperate with agent 0 first to open the switch, and then push the box together with agent 2. This role change of agent 1 brings new difficulty to the whole task. Agent 1 without options or with single-agent options tend to be stuck by the local optimum, after it turns on the switch with agent 0. However, agent 1 with multi-agent options (grouping: [0, 1], [1, 2]) can avoid this, since the adoption of the multi-agent option for the sub-group [1, 2] can encourage the exploration of the joint state space of agent 1 and 2.
}

\label{fig:7} 
\end{figure*}

\section{Evaluation} \label{eval}
Generally, a complex multi-agent task can be divided into some sub-tasks and each sub-task can be completed by a sub-group of the agents, and meanwhile, the sub-groups need to coordinate with each other very well to get the whole task done. In our algorithm, under a hierarchical MARL framework, we learn the optimal grouping through the attention mechanism, learn the sub-skill (option) of each sub-group through the multi-agent option discovery and learn the coordination among the sub-groups through the high-level policy learning. In this section, we will test the effectiveness of our algorithm on a self-designed simulator which is an extension of the commonly-used benchmark -- Push-Box \cite{lowe2017multi}.

\subsection{Simulator Setup}

As shown in Figure \ref{fig:5}, there are four agents in the environment and their task is to push the box to the target. During this process, they have to pass through a wall with a lock which is controlled by a switch, so they have to turn on the switch first to open the gate connecting the two rooms, and then push the box to the target. Note that only when two or more agents touch the switch at the same time will the switch open, and the box would not move until two or more agents push it in the same direction at the same time. In this case, the whole task can be divided into two sub-tasks, and coordination among the agents is required for each sub-task.

We compare our algorithm with two baselines: (1) a \textbf{SOTA} multi-agent reinforcement learning algorithm with attention mechanism but without option discovery -- MSAC \cite{DBLP:conf/icml/IqbalS19}; (2) a \textbf{SOTA} hierarchical multi-agent reinforcement learning algorithm with single-agent option discovery -- \textit{Deep Covering Option Discovery} \cite{DBLP:conf/iclr/JinnaiPMK20}.

The evaluation is based on the three sub-settings: (1) In this sub-setting, we simplify the task scenario to a two-agent version: agent 0 is able to turn on the switch, agent 1 is able to push the box, and both sub-tasks require only one agent. We will test the two baseline algorithms on this sub-setting and show the effectiveness of the single-agent option discovery method on a relatively simple task. (2) Only two of the agents (agent 0 and 1) are able to turn on the switch and the other two agents (agent 2 and 3) are able to push the box. We will see whether the grouping (group 0: agent 0 and 1, group 1: agent 2 and 3) can be found through the attention mechanism, and whether the sub-tasks (i.e., group 0: turning on the switch, group 1: pushing the box to the target) can be learned through our algorithm. (3) We only have three agents in this sub-setting: agent 0 is only able to turn on the switch, agent 2 is only able to push the box, while agent 1 can do both. In this case, agent 1 should first group with agent 0 to turn on the switch and then group with agent 2 to push the box to the target. For agent 1, the grouping is dynamic, which brings more challenges to this sub-setting.

Each agent takes its two-dimensional position (continuous) as the observation and chooses to execute one of the primitive actions in \{\textit{Up}, \textit{Down}, \textit{Left}, \textit{Right}, \textit{Stay}\} or the available options. Further, the reward function for each agent is defined as:
\begin{equation} \label{equ:19}
\begin{aligned}
    r^{i}_{env} = -c_{0} + c_{1}e^{i}_{switch} + c_{2}e^{i}_{move} + c_{3}e_{target}
\end{aligned}
\end{equation}
where $c_{0:3}>0$ are the weights for each term, $e^{i}_{move}$ represents whether the box is moved for one step by agent $i$, $e^{i}_{switch}$ represents whether the switch is turned on by agent $i$, and $e_{target}$ represents whether the box is pushed to the target area successfully (this term is shared by all the agents). Besides, the whole task is required to be completed within 400 steps, and a negative term $-c_{0}$ is used to encourage the agents to complete the whole task as soon as possible. Note that this reward setting is very sparse, since only when an event with low probability (e.g., reaching the target) occurs, can the agents get the corresponding reward term. Thus, highly-efficient exploration strategy is required. Baseline (1) is a \textbf{SOTA} MARL algorithm, but it fails in all the three sub-settings in our evaluation task (shown in Section \ref{result}), which shows that the evaluation scenarios are quite challenging.

\subsection{Results and Discussion} \label{result}

In this section, we compare MARL with no options (baseline (1)), single-agent options (baseline (2)), and multi-agent options (our algorithm) on the three sub-settings introduced above.

In the sub-setting 1, there are only two agents and each of them is responsible for a different task, i.e., agent 0: turning on the switch, agent 1: pushing the box to the target. Figure \ref{fig:6(a)}-\ref{fig:6(b)} plot the reward function of agent 0 and 1 in the training process. It can be observed that the agent without options can learn how to turn on the switch and get the corresponding reward term, however, it cannot learn to complete the harder task -- pushing the box to the target, due to the fact that the reward function is sparse. Then, we train a single-agent option for each agent at step 2000 with the transitions collected in the first 2000 steps, after which we start to train the high-level policy and low-level policy together within a hierarchical MARL framework introduced in Section \ref{framework}. The performance starts to improve at step 2000, which shows the effectiveness of MARL with the single-agent options. Agent 1 learns its sub-task through a denser reward function (Equation (\ref{equ:18})), and also agent 0 and 1 learns how to cooperate with each other (turn on the switch first and then push the box through the gate to the target) through the training of the high-level policy.

In the sub-setting 2, agent 0 and 1 need to touch the switch at the same time to turn it on, and agent 2 and 3 need to push the box in the same direction to make it move, so collaboration is required for each sub-task, which is different from the sub-setting 1. Figure \ref{fig:6(c)}-\ref{fig:6(d)} show that the agent without options or with four single-agent options (one for each agent) can learn to open the switch, while it cannot learn how to push the box to the target very well. Then, based on the transitions collected in the first 5000 steps, we can get the attention weight distribution of the agents. As shown in Figure \ref{fig:6(e)}, the attention weight that agent 0 pays to agent 1 is higher than the threshold 0.33 ($\approx 1/(4-1)$, as described Section \ref{network}) and vice versa, so we can get a sub-group: [0, 1]; similarly, we can get the other sub-group: [2, 3]. Then, we can train the multi-agent option for each sub-group to complete its corresponding sub-task, and integrate the options in the whole learning process as shown in Algorithm \ref{alg:Algo1}. Results show that the agent with multi-agent options performs significantly better, since it considers the collaboration among the agents in the option discovery process. While, the agent with single-agent options does not 
consider that, so it can do well in the sub-setting 1 but not this sub-setting.

Different from the sub-setting 2, in the sub-setting 3, agent 1 should cooperate with agent 0 first to open the switch, and then push the box to the target area together with agent 2, which makes the task even more challenging. Results in Figure \ref{fig:7} show that agent 1 without options or with single-agent options (grouping: [0], [1], [2]) tend to be stuck by the local optimum, after it completes the easier task -- opening the switch. While, agent 1 with multi-agent options can avoid this. The agents are divided into two sub-groups: [0, 1] and [1, 2] based on the attention weight distribution shown as Figure \ref{fig:6(f)} with Algorithm \ref{alg:Algo3} ($z$ is set as 0.5 $= 1/(3-1)$). The adoption of the multi-agent option for group [1, 2] can encourage the exploration of the joint state space of agent 1 and 2, and thus improve their overall performance. Note that the grouping and multi-agent option discovery are based on the transitions collected in the first 5000 steps, and the performance of agents with multi-agent options starts to improve after that.

\section{Conclusion} \label{conc}

This paper proposes \textit{Multi-agent Deep Covering Option Discovery} and a hierarchical multi-agent reinforcement learning algorithm based on soft actor-critic to integrate the options in a MARL setting -- \textbf{HA-MSAC}. This approach first divides all the agents into some sub-groups through the widely-used attention mechanism based on their interaction relationships, and then learns the multi-agent options for each sub-group to encourage the joint exploration of the multiple agents in a sub-group. Evaluation results show effective sub-group division through the attention mechanism, and superior performance of the MARL agents with multi-agent options as compared to the ones with single-agent options or no options. 

As a direction for future research, better algorithms for the sub-group division may be considered for more complicated collaborative tasks. Also, in this paper, the multi-agent option discovery is based on the joint observation space of the agents within a sub-group. If the size of a sub-group is large, its joint observation space can still be huge. Therefore, how to discover the multi-agent options based on the partial observation space of each individual agent is also an important future direction. Learning with options brings better interpretability and can potentially lead to explainable RL \cite{chen2022explain, chen2022relax}. As for application, options enable scheduling with multiple levels of temporal abstractions and can benefit intelligent transportation research \cite{ma2020statistical, luo2022multisource}. 


\bibliography{main}
\bibliographystyle{IEEEtran}

\end{document}